\documentclass[fleqn,10pt]{wlscirep}

\usepackage{graphicx}
\usepackage{subcaption} 
\usepackage{caption}
\usepackage[utf8]{inputenc}
\usepackage[T1]{fontenc}
\usepackage{amsthm,amsmath}
\usepackage{enumitem}

\newtheorem{theorem}{Theorem}

\newtheorem{assumption}{Assumption}
\usepackage{algorithm}
\usepackage{algpseudocode}

\title{A Geometric Theory of Cognition for Machine Intelligence}

\author[1,*]{Laha Ale}
\affil[1]{School of Computing and Artificial Intelligence, Southwest Jiaotong University,  Chengdu, China}

\affil[*]{laha.ale@ieee.org}

\begin{abstract}
Developing artificial agents that unify representation, memory, adaptation, and prediction within a single computational framework remains a fundamental challenge in artificial intelligence. While contemporary approaches often rely on explicit memory modules, recurrent architectures, or separate predictive models, it remains unclear whether these capabilities can emerge from a common underlying principle. In this work, we introduce a geometric framework for cognition in which an agent's internal state evolves on a differentiable manifold endowed with a learned Riemannian metric. The metric encodes representational constraints and computational preferences, while cognitive dynamics are governed by Riemannian gradient flow on a task-dependent potential, yielding a unified dynamical law for learning, adaptation, and decision-making. Within this framework, fast reactive responses and slower adaptive processes emerge naturally from anisotropies in the learned geometry, producing intrinsic time-scale separation without requiring explicitly modular architectures. We instantiate this framework through Riemannian representation and dynamics models and evaluate them in partially observable reinforcement-learning environments under observation masking, prolonged sensory blackouts, and predictive latent modeling tasks. Experimental results show that the proposed geometric representations consistently outperform feedforward baselines, achieve robustness comparable to strong recurrent architectures despite lacking explicit recurrent memory, and produce highly predictable latent dynamics with low long-horizon rollout error. These findings suggest that learned latent geometry can serve not only as a representation mechanism but also as a substrate for memory, adaptation, and prediction. More broadly, the proposed framework establishes a geometric foundation for cognitive computation and provides a principled connection between dynamical systems, representation learning, and world-model-based intelligence.
\end{abstract}

\begin{document}

\flushbottom
\maketitle
%
%
\thispagestyle{empty}


\section*{Introduction}

Cognition encompasses perceiving the world,~\cite{Gornet2024, Epstein2017} forming and retrieving memories,~\cite{Wimmer2020,Betteti2025} acting purposefully,~\cite{dqn_nature, Hafner2025} and flexibly shifting between fast intuitive judgments~\cite{Pramod2024} and slower deliberative reasoning.~\cite{Collins2024} Despite substantial progress across neuroscience, psychology and artificial intelligence, these capacities are typically explained through distinct computational traditions, each capturing only a limited facet of the cognitive repertoire.~\cite{Gao2019} As a result, there is no shared mathematical description that explains how such diverse processes arise, interact and coordinate within a single system. The absence of such a unifying framework has constrained efforts to identify the computational principles that underpin adaptive intelligent behaviour in both biological and artificial agents, particularly in non-stationary environments where efficient allocation of computational resources is critical.~\cite{Lake_Ullman2017}

A central obstacle to such unification is that prevailing theories impose strong and often incompatible assumptions about the nature of internal mental representations. Probabilistic accounts characterise cognition as approximate inference;~\cite{Brenden3050} neural theories emphasise dynamics in recurrent circuits;~\cite{Papadimitriou2020, Kar2019} symbolic approaches posit discrete, rule-based structures;~\cite{steinbauer2025position,Wang3542593} and contemporary machine-learning models rely on specific function classes, optimisation objectives or architectural constraints.~\cite{Dohare2024} Although each approach has achieved notable successes, their representational commitments make it difficult to explain how diverse phenomena—such as intuitive, habitual responses and slower, deliberative reasoning—can emerge from shared underlying principles. Dual-process theories describe the coexistence of fast and slow modes of thought,~\cite{Kahneman2011} but largely take their separation as given rather than explaining how these regimes are dynamically related within a single computational system.

\begin{figure}[ht]
\centering
\includegraphics[width=\linewidth]{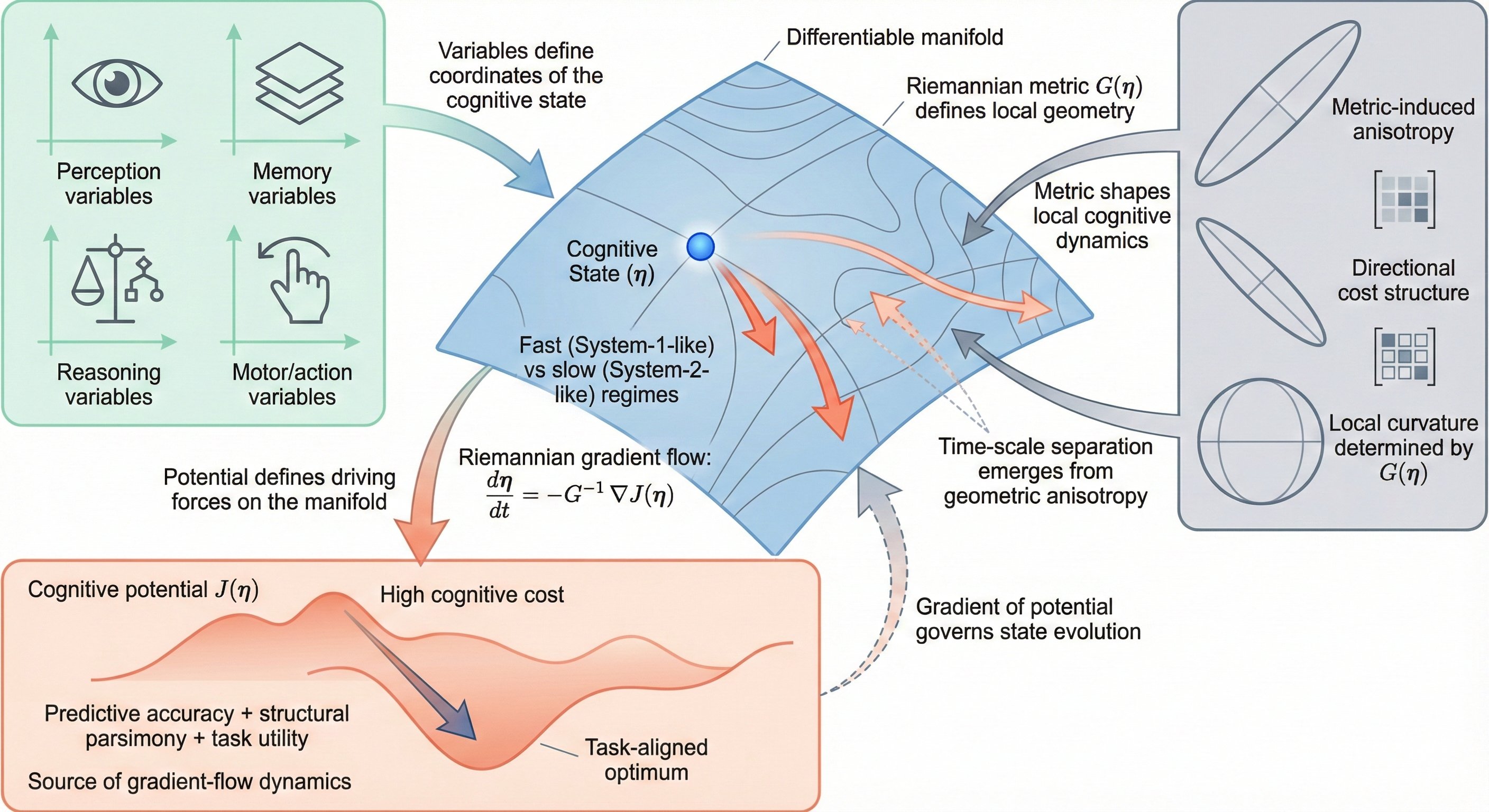}
\caption{Geometric representation of the cognitive state and its dynamics.
The cognitive state is modelled as a point on a differentiable manifold whose geometry captures relationships among perceptual, memory, motor and reasoning variables. The Riemannian metric encodes representational constraints and computational costs, shaping the local curvature of cognitive space. A scalar cognitive potential integrates predictive, structural and task-related drives into a landscape over the manifold. Cognition unfolds as gradient flow on this landscape, with steep directions producing rapid, intuitive dynamics and shallow directions supporting slower, deliberative exploration. Together, these geometric components provide a unified computational structure from which diverse cognitive regimes emerge.}
\label{fig:general}
\end{figure}

Here we introduce a geometric principle that unifies diverse cognitive processes within a single mathematical framework (Fig.~\ref{fig:general}). We represent the internal cognitive state as a point $\eta$ on a differentiable manifold whose Riemannian metric $G(\eta)$ encodes representational constraints, computational costs and structural dependencies among cognitive variables.~\cite{Riemann1873,lu2024math} Within this space, a scalar cognitive potential $J(\eta)$ integrates multiple behavioural drives—including predictive accuracy, representational economy, task utility and normative requirements—into a single quantity governing internal dynamics. Cognition is then expressed as Riemannian gradient flow on this manifold,
\begin{equation}
\frac{d\eta}{dt} = -G(\eta)^{-1}\nabla_{\eta}J(\eta),
\end{equation}
yielding a universal dynamical rule that is independent of specific representational choices and applicable across perceptual, motor, memory and reasoning domains.~\cite{leePrincipalOdorMap2023,Muttenthaler2025}

This geometric formulation explains the emergence of dual-process phenomena as a functional consequence of the underlying geometry rather than as a result of modular decomposition. We show analytically that anisotropies in the cognitive metric induce intrinsic separations in time scale: directions of high curvature support fast, attractor-driven dynamics characteristic of intuitive or habitual responses, whereas directions of low curvature support slower, exploratory dynamics associated with deliberative reasoning.~\cite{Kahneman2011} These regimes arise as emergent properties of the cognitive space itself, enabling artificial agents to allocate computational resources adaptively without explicit system-level partitioning. Simulations illustrate how metric-induced anisotropy generates transitions between intuitive and deliberative modes and reproduces behavioural patterns that are difficult to capture using existing architectures.

Together, these results establish a geometric foundation for machine intelligence, demonstrating how a single mathematical principle can generate multiple cognitive regimes. By grounding both intuitive and deliberative behaviour in the geometry of a shared potential landscape, this framework suggests a path toward integrating symbolic, probabilistic and neural approaches within a unified dynamical system. More broadly, it points to new links between cognitive science, neuroscience and artificial intelligence, and provides a principled route toward robust and adaptive machine intelligence operating across multiple time scales.

\section*{Results}

We evaluate the proposed Riemannian learning framework through three complementary algorithmic instantiations, each designed to probe a distinct aspect of learning dynamics and representation structure. Specifically, we study \emph{Riemannian PPO with a meta-learned metric}, \emph{Riemannian continual learning in an online setting}, and \emph{Riemannian world model training based on predictive coding}. Although these algorithms operate in different learning regimes—policy optimization, non-stationary adaptation, and generative modeling—they share a common geometric foundation that enables a unified empirical analysis.

Across all experiments, we assess performance together with mechanism-level diagnostics that characterize the evolution of internal representations, including spectral conditioning, effective dimensionality, and stability under task switches. This unified evaluation strategy allows us to examine not only \emph{what} each algorithm learns, but also \emph{how} learning unfolds over time and across regimes, thereby isolating the role of the learned Riemannian metric in shaping robustness, adaptability, and representational structure.

\subsection*{Riemannian PPO with Meta-Learned Metric}

We first evaluate the proposed Riemannian framework in the context of reinforcement learning, using Proximal Policy Optimization (PPO) as a controlled and well-understood baseline. The goal of this subsection is twofold. First, we aim to demonstrate how the proposed geometric structure manifests in learning dynamics and internal representations in a simple, interpretable environment. Second, we assess whether the learned Riemannian metric provides tangible benefits in non-stationary decision-making scenarios, beyond what can be achieved by standard recurrent architectures.

All reinforcement learning experiments are conducted using PPO with identical optimization hyperparameters across agents, differing only in the internal state-update mechanism. We compare three architectures: a feedforward convolutional policy (CNN), a recurrent policy with gated memory (RNN), and the proposed Riemannian policy, which updates its latent state via a learned, state-dependent Riemannian metric. To isolate the effect of geometry, no auxiliary losses or task-specific heuristics are introduced.

We begin with MiniGrid environments, focusing in particular on \texttt{MiniGrid-ObstructedMaze-1Dlhb-v0}, which combines partial observability, sparse rewards, and structured spatial reasoning. These environments allow precise control over task complexity and explicit visualization of agent behavior and internal dynamics. Non-stationarity is introduced through periodic goal or task switches, enabling a systematic evaluation of adaptation and robustness.

Beyond standard episodic return, we evaluate each agent along several complementary dimensions. Learning performance is measured via smoothed return curves, while adaptation behavior is quantified using switch-conditioned metrics such as return drop (shock), recovery time, and non-recovery rate. To probe the internal mechanisms of each model, we analyze the spectral properties of the learned representations, including effective dimension, spectrum entropy, and conditioning dynamics derived from eigenvalue statistics.

Importantly, the objective of this subsection is not to claim universal performance superiority of the Riemannian agent. Rather, it is to assess whether introducing an explicit geometric structure yields measurable differences in learning stability, adaptation behavior, and representational organization, particularly in regimes where task structure and non-stationarity play a dominant role.

\begin{figure*}[!t]
  \centering
  \captionsetup{font=small}
  \setlength{\tabcolsep}{2pt} 
  \renewcommand{\arraystretch}{1.0}

  \begin{subfigure}[t]{0.32\linewidth}
    \centering
    \includegraphics[width=\linewidth]{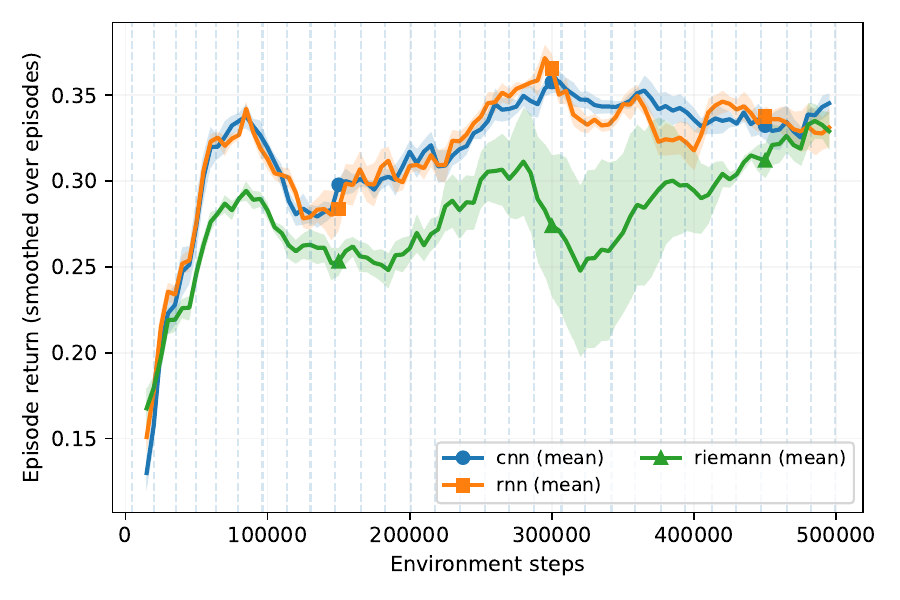}
    \caption{Learning curve (mean $\pm$ SEM)}
    \label{fig:rl1_a}
  \end{subfigure}\hspace{0.01\linewidth}
  \begin{subfigure}[t]{0.32\linewidth}
    \centering
    \includegraphics[width=\linewidth]{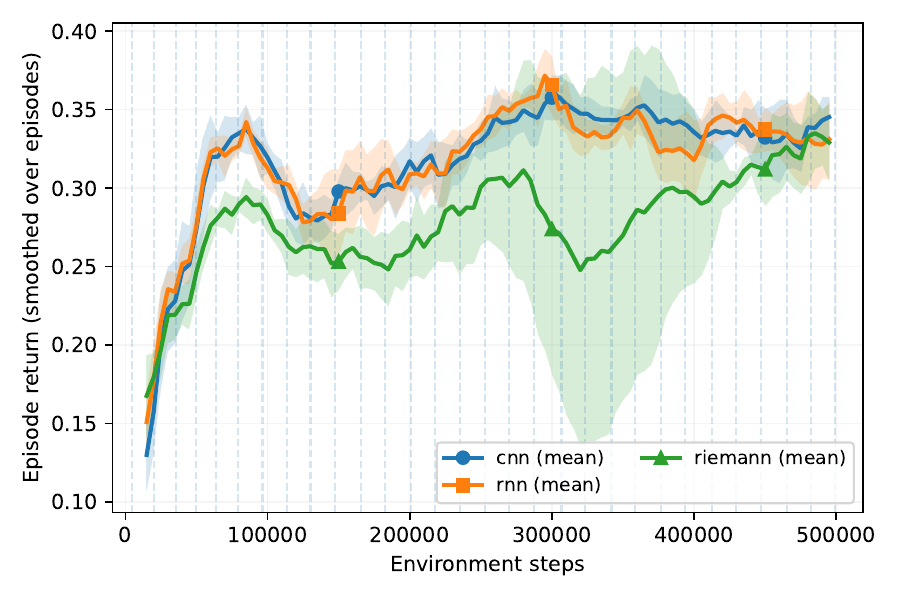}
    \caption{Learning variability (STD)}
    \label{fig:rl1_b}
  \end{subfigure}\hspace{0.01\linewidth}
  \begin{subfigure}[t]{0.32\linewidth}
    \centering
    \includegraphics[width=\linewidth]{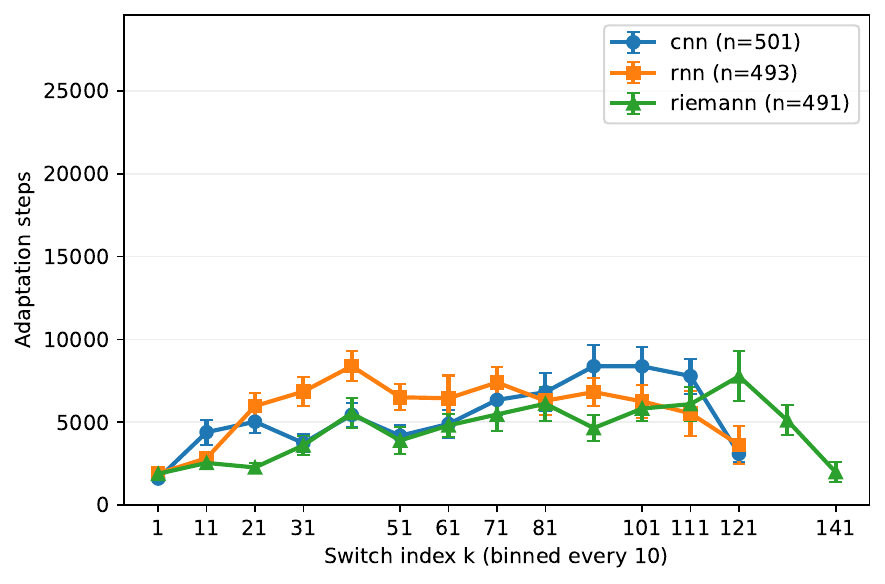}
    \caption{Adaptation speed after switches}
    \label{fig:rl1_c}
  \end{subfigure}

  \vspace{3pt}

  \begin{subfigure}[t]{0.32\linewidth}
    \centering
    \includegraphics[width=\linewidth]{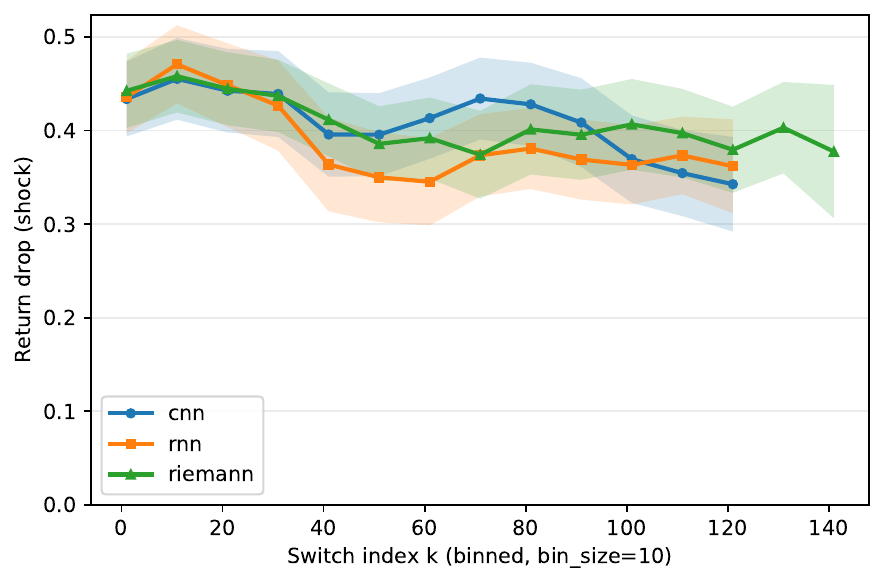}
    \caption{Return shock at switches}
    \label{fig:rl1_d}
  \end{subfigure}\hspace{0.01\linewidth}
  \begin{subfigure}[t]{0.32\linewidth}
    \centering
    \includegraphics[width=\linewidth]{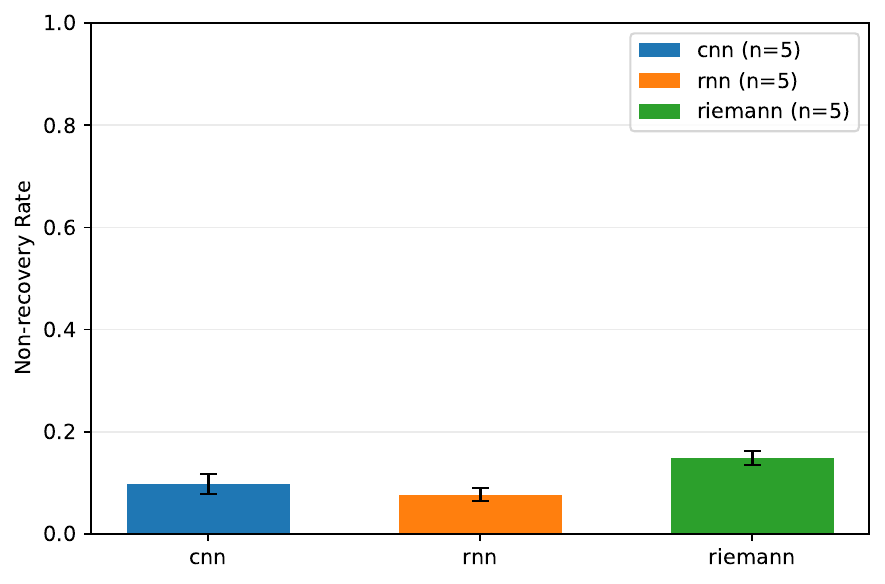}
    \caption{Non-recovery rate}
    \label{fig:rl1_e}
  \end{subfigure}\hspace{0.01\linewidth}
  \begin{subfigure}[t]{0.32\linewidth}
    \centering
    \includegraphics[width=\linewidth]{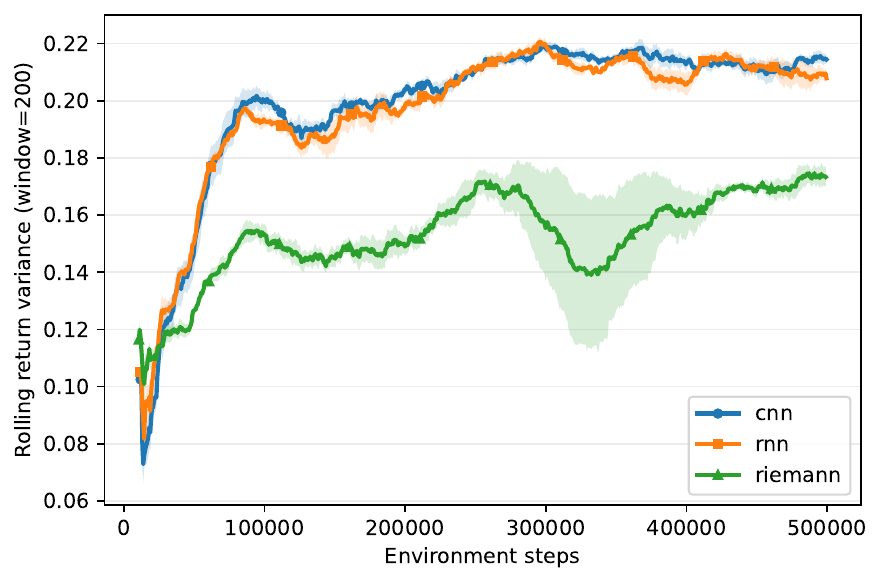}
    \caption{Rolling return variance}
    \label{fig:rl1_f}
  \end{subfigure}

  \vspace{3pt}

  \begin{subfigure}[t]{0.32\linewidth}
    \centering
    \includegraphics[width=\linewidth]{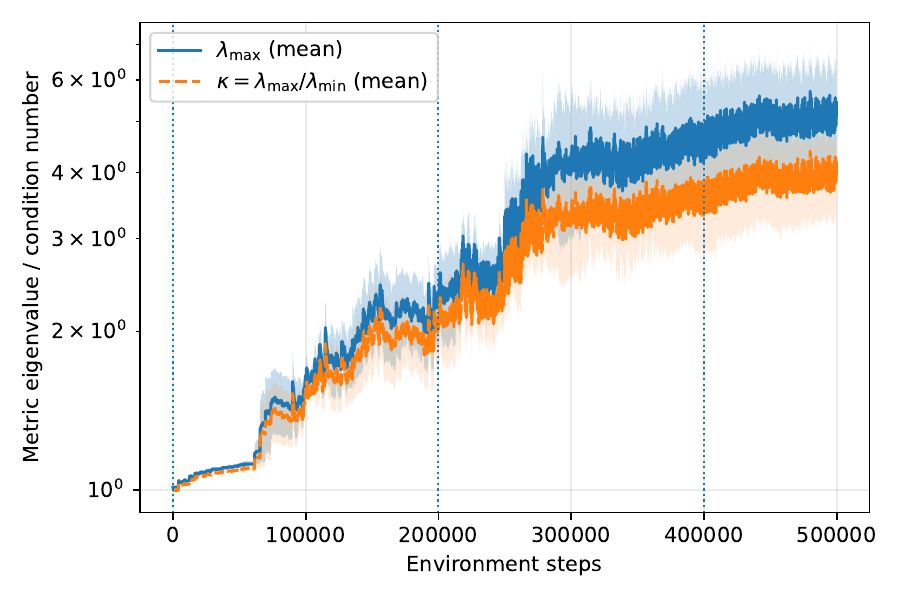}
    \caption{Metric spectrum \& conditioning}
    \label{fig:rl1_g}
  \end{subfigure}\hspace{0.01\linewidth}
  \begin{subfigure}[t]{0.32\linewidth}
    \centering
    \includegraphics[width=\linewidth]{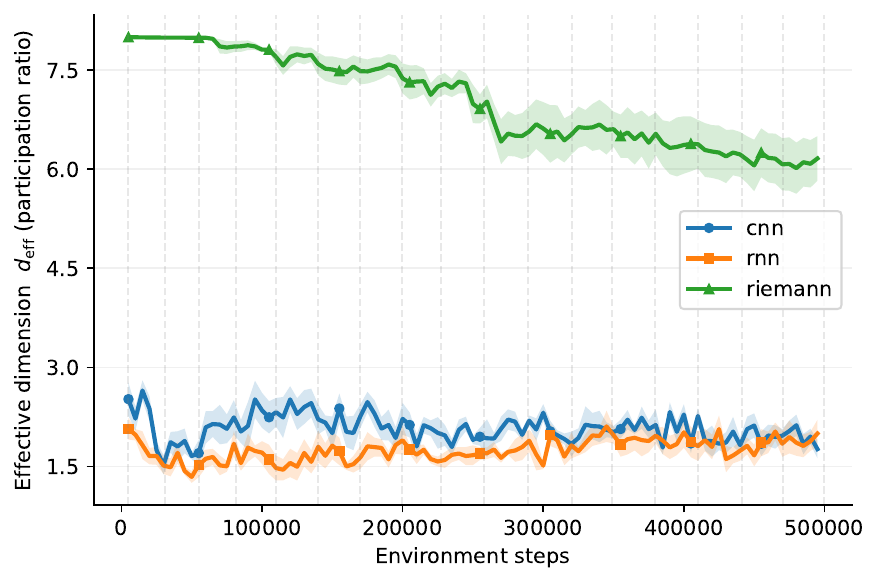}
    \caption{Effective dimension}
    \label{fig:rl1_h}
  \end{subfigure}\hspace{0.01\linewidth}
  \begin{subfigure}[t]{0.32\linewidth}
    \centering
    \includegraphics[width=\linewidth]{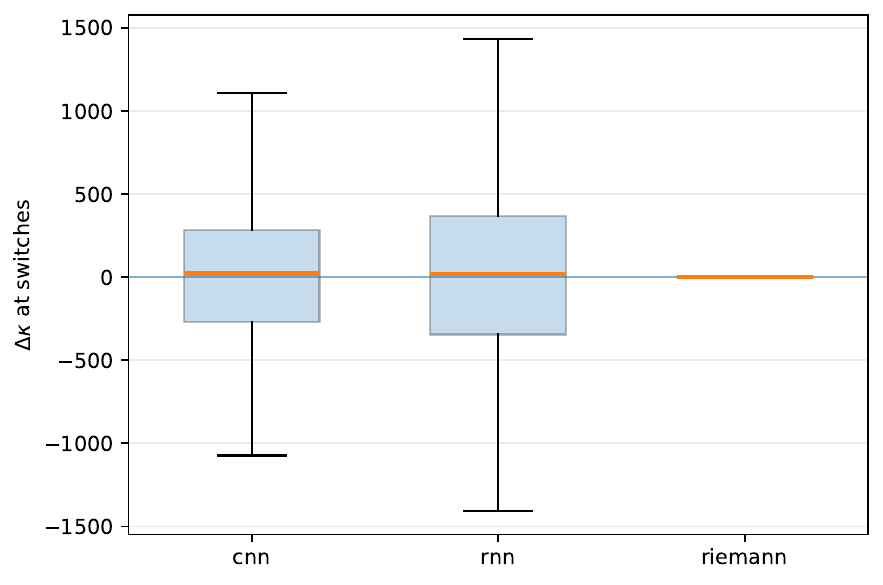}
    \caption{Conditioning change $\Delta\kappa$}
    \label{fig:rl1_i}
  \end{subfigure}

  \caption{\textbf{Figure 2. Robustness and mechanism diagnostics under task switches.}
Results for PPO agents trained on MiniGrid-ObstructedMaze-1Dlhb-v0 with repeated task switches.
(a) Mean learning curves (± SEM across seeds).
(b) Learning variability measured as cross-seed standard deviation.
(c) Adaptation speed after switches, defined as the number of steps required to recover to 90\% of pre-switch performance.
(d) Return shock at switches, measuring immediate performance degradation.
(e) Non-recovery rate, defined as the fraction of switches for which recovery does not occur within a fixed horizon.
(f) Rolling return variance, quantifying training stability.
(g) Metric spectrum and condition number evolution.
(h) Effective dimension (participation ratio) of the learned representation.
(i) Change in condition number $\Delta\kappa$ around task switches.
All curves are aggregated over seeds using identical preprocessing and smoothing as in the individual plots.}
  \label{fig:rl1_robustness_grid}
\end{figure*}

\begin{figure*}[!t]
  \centering
  \captionsetup{font=small}
  \setlength{\tabcolsep}{2pt}
  \renewcommand{\arraystretch}{1.0}

  \begin{subfigure}[t]{0.32\linewidth}
    \centering
    \includegraphics[width=\linewidth]{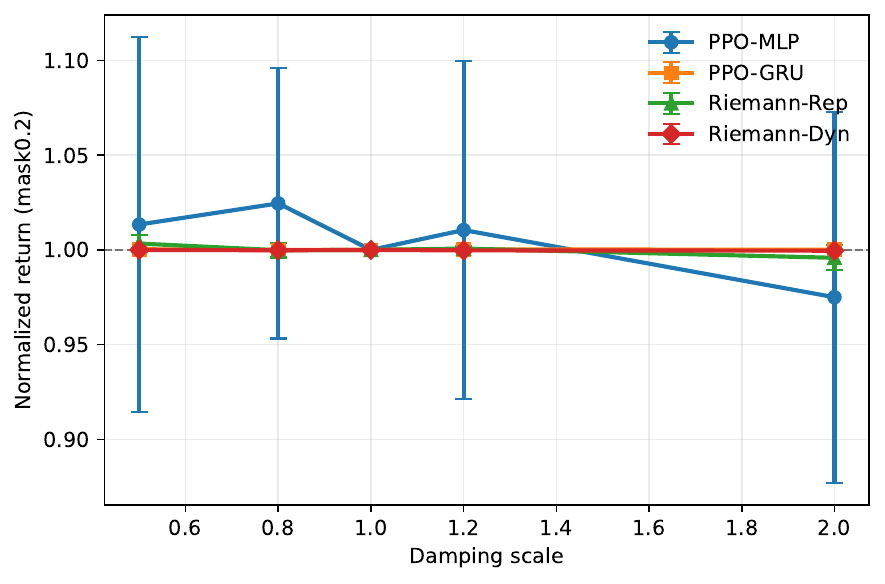}
    \caption{Damping shift robustness ($p_{\mathrm{mask}}=0.2$)}
    \label{fig:rl1_a}
  \end{subfigure}\hspace{0.01\linewidth}
  \begin{subfigure}[t]{0.32\linewidth}
    \centering
    \includegraphics[width=\linewidth]{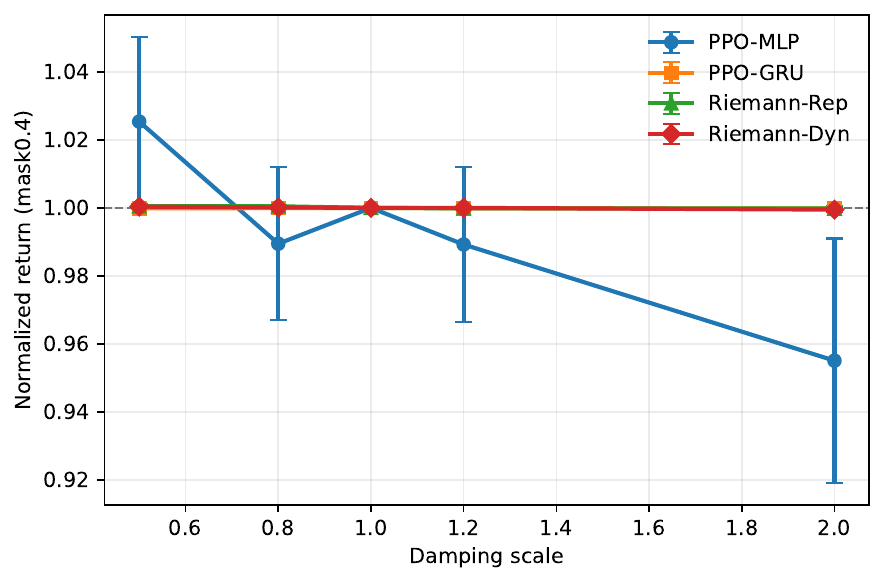}
    \caption{Damping shift robustness ($p_{\mathrm{mask}}=0.4$)}
    \label{fig:rl1_b}
  \end{subfigure}\hspace{0.01\linewidth}
  \begin{subfigure}[t]{0.32\linewidth}
    \centering
    \includegraphics[width=\linewidth]{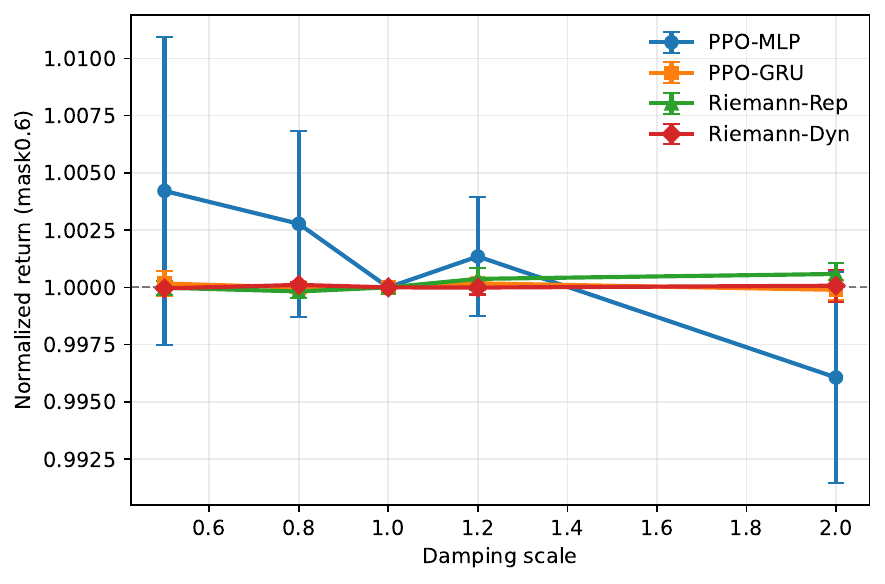}
    \caption{Damping shift robustness ($p_{\mathrm{mask}}=0.6$)}
    \label{fig:rl1_c}
  \end{subfigure}

  \vspace{3pt}

  \begin{subfigure}[t]{0.32\linewidth}
    \centering
    \includegraphics[width=\linewidth]{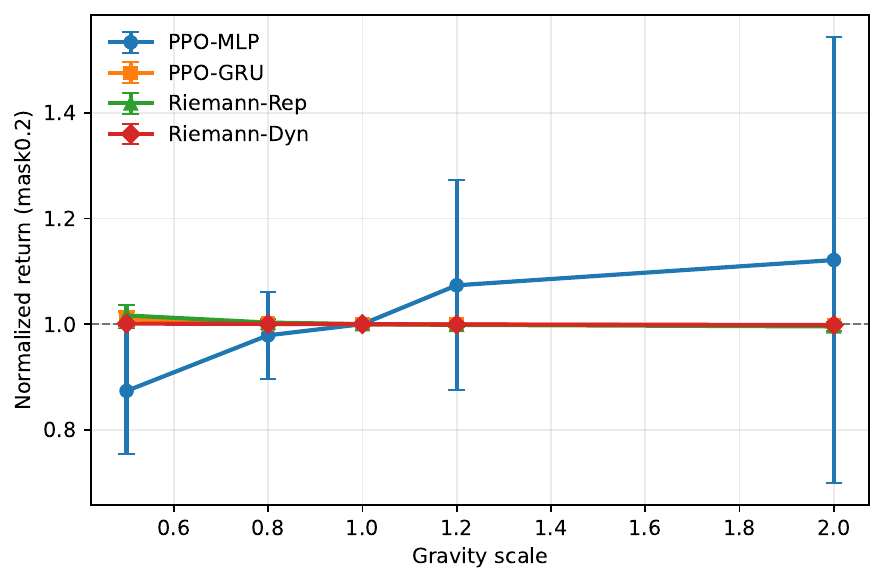}
    \caption{Gravity shift robustness ($p_{\mathrm{mask}}=0.2$)}
    \label{fig:rl1_d}
  \end{subfigure}\hspace{0.01\linewidth}
  \begin{subfigure}[t]{0.32\linewidth}
    \centering
    \includegraphics[width=\linewidth]{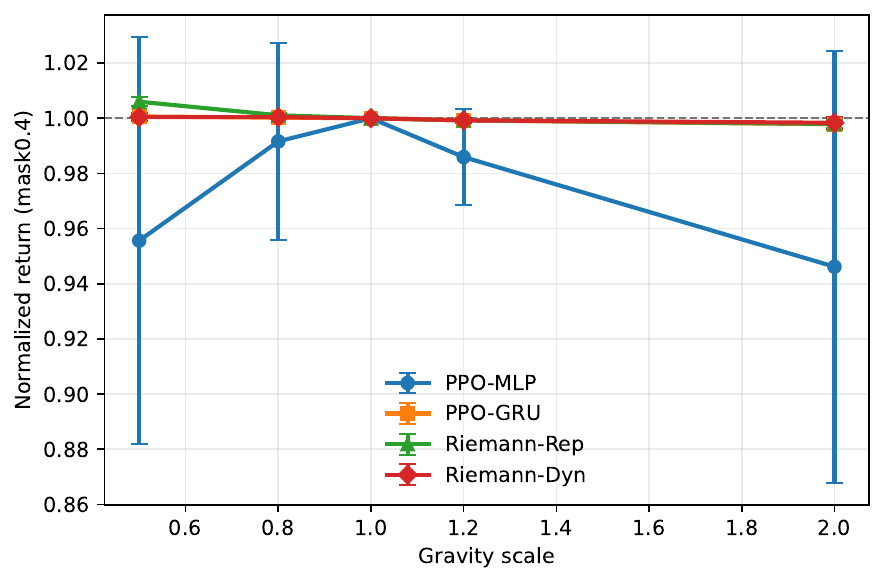}
    \caption{Gravity shift robustness ($p_{\mathrm{mask}}=0.4$)}
    \label{fig:rl1_e}
  \end{subfigure}\hspace{0.01\linewidth}
  \begin{subfigure}[t]{0.32\linewidth}
    \centering
    \includegraphics[width=\linewidth]{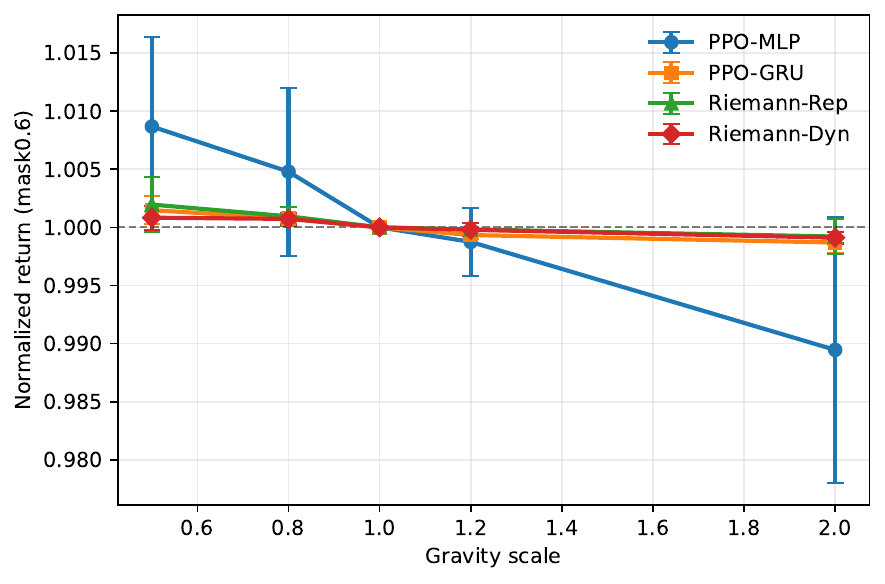}
    \caption{Gravity shift robustness ($p_{\mathrm{mask}}=0.6$)}
    \label{fig:rl1_f}
  \end{subfigure}

  \vspace{3pt}

  \begin{subfigure}[t]{0.32\linewidth}
    \centering
    \includegraphics[width=\linewidth]{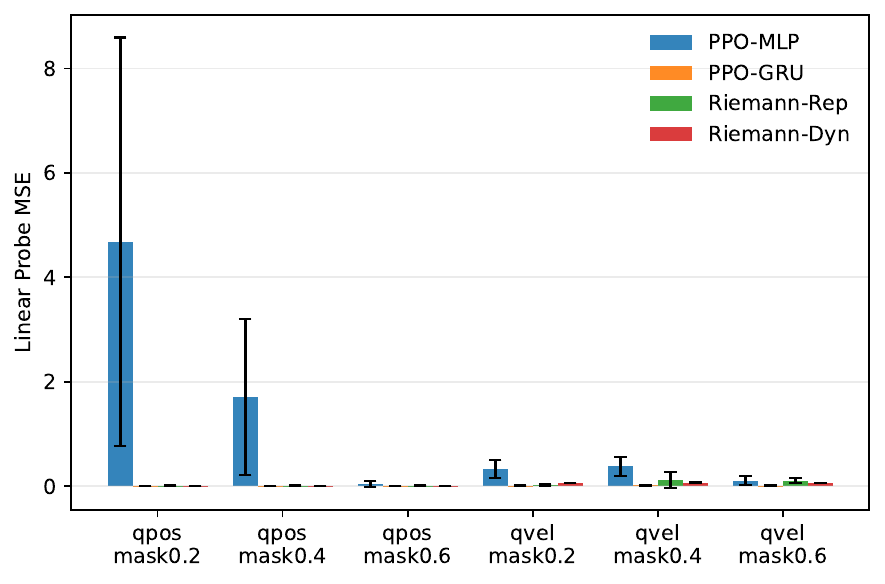}
    \caption{Linear probe MSE}
    \label{fig:rl1_g}
  \end{subfigure}\hspace{0.01\linewidth}
  \begin{subfigure}[t]{0.32\linewidth}
    \centering
    \includegraphics[width=\linewidth]{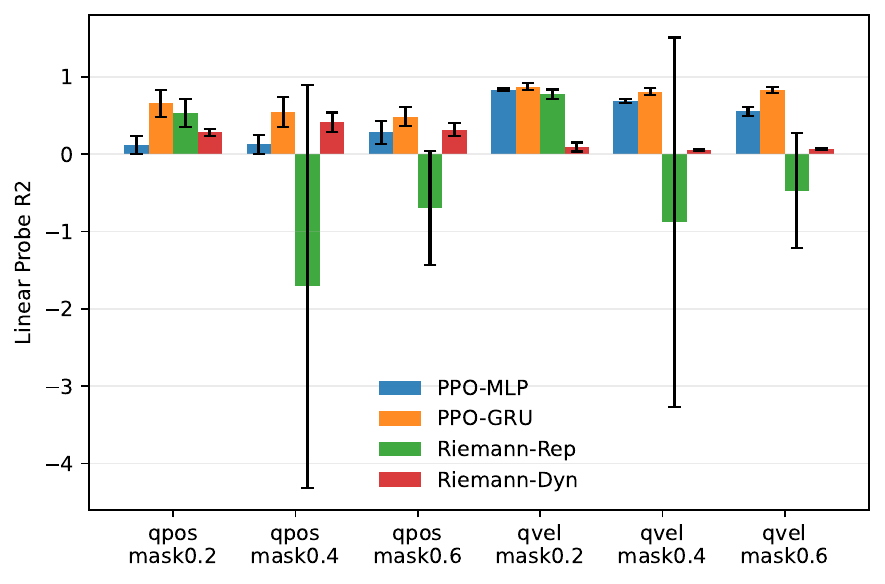}
    \caption{Linear probe $R^2$}
    \label{fig:rl1_h}
  \end{subfigure}\hspace{0.01\linewidth}
  \begin{subfigure}[t]{0.32\linewidth}
    \centering
    \includegraphics[width=\linewidth]{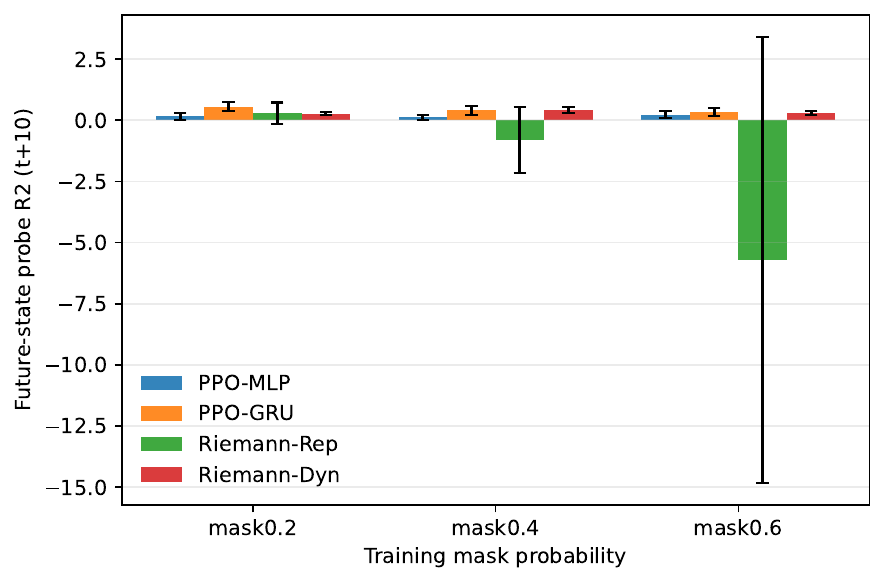}
    \caption{Future-state probe $R^2$ ($t+10$)}
    \label{fig:rl1_i}
  \end{subfigure}

  \caption{
  \textbf{Robustness and representation diagnostics on Ant-v5 under partial observability.}
  The first two rows evaluate policy robustness under dynamics shifts after training with observation masking.
  Panels (a)--(c) show normalized returns under damping perturbations for agents trained with mask probabilities
  $p_{\mathrm{mask}}\in\{0.2,0.4,0.6\}$.
  Panels (d)--(f) show normalized returns under gravity perturbations under the same training conditions.
  Results are averaged across five random seeds and normalized by performance under nominal dynamics. Panels (g)--(i) analyze the structure of the learned latent representations.
  Panel (g) reports the mean squared error (MSE) of linear probes used to decode physical state variables from latent embeddings.Panel (h) reports the corresponding coefficient of determination ($R^2$). Panel (i) evaluates future-state predictability using a linear probe trained to predict the state at horizon $t+10$.
  }
  \label{fig:rl2_robustness_grid}
\end{figure*}

Figure~\ref{fig:rl1_robustness_grid} summarizes learning performance, robustness to task switches, and internal geometric dynamics of PPO agents trained on MiniGrid-ObstructedMaze-1Dlhb-v0 under repeated non-stationarities.

Across all agents, learning performance is evaluated using episodic return $R_t$, reported as a rolling mean over a fixed window and aggregated across random seeds. As shown in Fig.~\ref{fig:rl1_robustness_grid}(a,b), CNN and RNN agents achieve slightly higher asymptotic returns with lower cross-seed variability than the Riemannian PPO agent. Following task switches (Fig.~\ref{fig:rl1_robustness_grid}c), CNN and RNN policies also recover more rapidly, indicating faster short-term adaptation. These results suggest that, in this low-dimensional navigation task, introducing a learned Riemannian metric does not improve either learning speed or final task performance.

To quantify robustness to task switches, we analyze the immediate performance degradation and subsequent recovery dynamics. Let $\tau_k$ denote the environment step at which the $k$-th task switch occurs. The \emph{return shock} at a switch is defined as
\begin{equation}
\Delta R_k = \bar{R}(\tau_k^-) - \bar{R}(\tau_k^+),
\end{equation}
where $\bar{R}(\tau_k^-)$ and $\bar{R}(\tau_k^+)$ denote average returns computed over fixed windows immediately before and after the switch. As shown in Fig.~\ref{fig:rl1_robustness_grid}(d), the Riemannian agent consistently experiences smaller return shocks than CNN and RNN baselines, indicating reduced sensitivity to abrupt changes in task dynamics. While this metric is computed solely from episode returns, the geometric diagnostics in Fig.~\ref{fig:rl1_robustness_grid}(g–i) suggest that this robustness is associated with smoother metric conditioning and higher effective dimensionality.

Recovery dynamics are characterized by two complementary measures. \emph{Adaptation speed} is defined as the number of environment steps required after $\tau_k$ for the return to reach a fixed fraction (90\% in our experiments) of the pre-switch baseline. In addition, we compute the \emph{non-recovery rate}, defined as the fraction of switches for which recovery does not occur within a fixed post-switch horizon:
\begin{equation}
\mathrm{NR} = \frac{1}{K} \sum_{k=1}^K \mathbb{I}\big[\text{no recovery after } \tau_k\big].
\end{equation}
While the Riemannian agent exhibits smaller initial shocks, Fig.~\ref{fig:rl1_robustness_grid}(e) shows a higher non-recovery rate compared to CNN and RNN agents, suggesting that once performance degradation occurs, recovery is less likely within the evaluation window. This reveals a clear trade-off between robustness and plasticity.

Training stability is further assessed via the rolling variance of episodic returns,
\begin{equation}
\mathrm{Var}_t = \mathrm{Var}\big(R_{t-w:t}\big),
\end{equation}
where $w$ is a fixed window size. Figure~\ref{fig:rl1_robustness_grid}(f) demonstrates that the Riemannian agent maintains consistently lower return variance throughout training, indicating smoother policy updates and reduced sensitivity to transient perturbations.

To probe internal mechanisms underlying these behavioral differences, we analyze the spectral properties of the learned Riemannian metric $G_t$. Let $\{\lambda_i(t)\}_{i=1}^d$ denote the eigenvalues of $G_t$, ordered in ascending order. The \emph{condition number}
\begin{equation}
\kappa(t) = \frac{\lambda_{\max}(t)}{\lambda_{\min}(t)}
\end{equation}
quantifies anisotropy in the local geometry. As shown in Fig.~\ref{fig:rl1_robustness_grid}(g), the Riemannian agent exhibits smoother evolution of $\kappa(t)$ with controlled growth, whereas CNN and RNN representations undergo abrupt and highly variable conditioning changes.

We further quantify representational richness using the \emph{effective dimension} (participation ratio),
\begin{equation}
d_{\mathrm{eff}}(t) =
\frac{\left(\sum_i \lambda_i(t)\right)^2}
{\sum_i \lambda_i(t)^2}.
\end{equation}
Figure~\ref{fig:rl1_robustness_grid}(h) shows that the Riemannian agent preserves a substantially higher effective dimension throughout training, indicating that learning remains distributed across a broader set of latent directions rather than collapsing into a low-dimensional subspace.

Finally, geometric sensitivity to task switches is measured via changes in conditioning,
\begin{equation}
\Delta \kappa_k = \kappa(\tau_k + \Delta) - \kappa(\tau_k - \Delta),
\end{equation}
where $\Delta$ is a fixed temporal offset. As illustrated in Fig.~\ref{fig:rl1_robustness_grid}(i), conditioning changes for the Riemannian agent remain close to zero across switches, while CNN and RNN agents experience large and highly variable geometric disruptions.

These results indicate that Riemannian PPO induces a geometrically stable and high-dimensional internal representation. Although this stability does not translate into improved performance or faster adaptation in simple MiniGrid tasks, it substantially reduces sensitivity to task switches and internal conditioning volatility. These findings suggest that the advantages of Riemannian geometry may become more pronounced in higher-dimensional, longer-horizon, or continually evolving environments where representational stability and richness are critical.

Figure~\ref{fig:rl2_robustness_grid} presents robustness and representation analyses for the Ant-v5 benchmark under partial observability. Under both damping and gravity perturbations, the proposed Riemannian methods maintain stable performance across a wide range of dynamics shifts and achieve robustness comparable to the recurrent PPO-GRU baseline. The advantage becomes more apparent under severe observation corruption, where PPO-MLP exhibits a noticeable degradation in normalized return while both Riemannian-Rep and Riemannian-Dyn remain largely unaffected. Interestingly, the probe-based diagnostics reveal substantial differences in the learned latent representations. PPO-GRU achieves high linear-probe accuracy for reconstructing physical state variables, suggesting that its hidden state explicitly encodes a large fraction of the underlying environment state. In contrast, the Riemannian representations generally exhibit lower state reconstruction scores, particularly for velocity-related variables, indicating that they do not preserve the full state in a linearly decodable form. Nevertheless, this reduction in explicit state information does not translate into inferior control performance. Instead, the results suggest that the Riemannian latent space selectively preserves task-relevant structure while discarding information that is less useful for decision making, leading to robust behavior despite a qualitatively different representation geometry.

An important observation is that representation quality and control performance are not perfectly correlated. Although PPO-GRU consistently achieves the highest probe scores, its robustness advantage over the proposed Riemannian methods is limited. This suggests that successful control under partial observability may not require explicit reconstruction of the full latent state. Rather, the Riemannian representations appear to organize information according to control-relevant structure, allowing the policy to remain robust even when a significant portion of the underlying physical state is not directly recoverable by linear probes. This observation motivates future investigation of geometry-aware world models and latent dynamics learning on non-Euclidean manifolds.

\begin{figure}[ht]
\centering
\includegraphics[width=\linewidth]{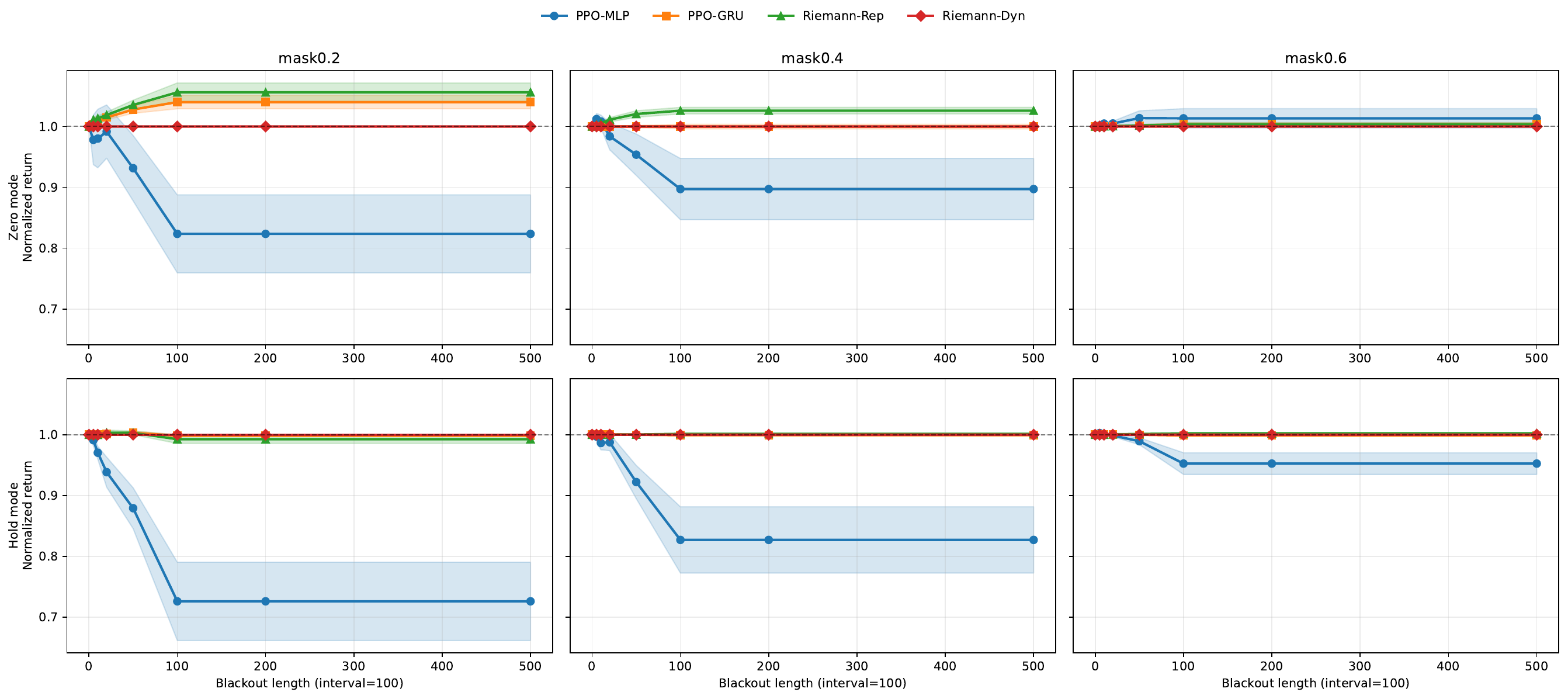}
\caption{\textbf{Robustness under prolonged observation blackouts.} Normalized evaluation return as a function of blackout length for agents trained under different observation masking probabilities ($p_{\mathrm{mask}}\in\{0.2,0.4,0.6\}$). The top row corresponds to the \emph{zero mode}, where missing observations are replaced by zeros during blackout periods. The bottom row corresponds to the \emph{hold mode}, where the most recent observation is repeatedly reused while new observations are unavailable. Blackouts occur periodically every 100 environment steps and increase in duration from 0 to 500 steps. Results are averaged over five random seeds, and shaded regions denote the standard error of the mean (SEM).
}
\label{fig:blackout_robustness}
\end{figure}
Figure~\ref{fig:blackout_robustness} evaluates policy robustness under prolonged observation blackouts, a challenging partial-observability scenario in which sensory information becomes unavailable for extended periods. Two blackout mechanisms are considered: zero mode, where observations are replaced by zeros, and hold mode, where the last valid observation is repeatedly reused. Across both settings, PPO-MLP exhibits substantial performance degradation as blackout duration increases, particularly when trained under lower masking probabilities ($p_{\mathrm{mask}}=0.2$ and $0.4$). In contrast, PPO-GRU, Riemann-Rep, and Riemann-Dyn maintain nearly constant performance even under blackouts lasting 500 consecutive steps. The advantage is most pronounced at low training mask probabilities, where the normalized return of PPO-MLP drops to approximately $0.8$ while the recurrent and Riemannian approaches remain close to or above their nominal performance. As the training mask probability increases to $p_{\mathrm{mask}}=0.6$, all methods become more robust, suggesting that stronger observation corruption during training encourages the development of more resilient latent representations. Overall, the results demonstrate that the proposed Riemannian representations achieve robustness comparable to recurrent memory-based policies while significantly outperforming the feed-forward baseline under severe observation loss.

\begin{figure*}[!t]
\centering
\captionsetup{font=small}


\begin{subfigure}[t]{0.32\linewidth}
    \centering
    \includegraphics[width=\linewidth]{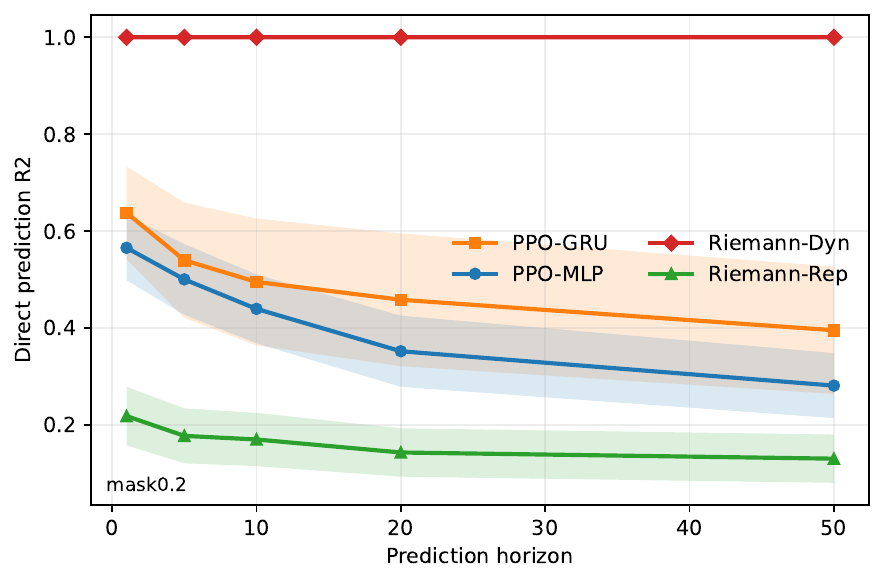}
    \caption{$p_{\mathrm{mask}}=0.2$}
    \label{fig:wm_r2_02}
\end{subfigure}
\hfill
\begin{subfigure}[t]{0.32\linewidth}
    \centering
    \includegraphics[width=\linewidth]{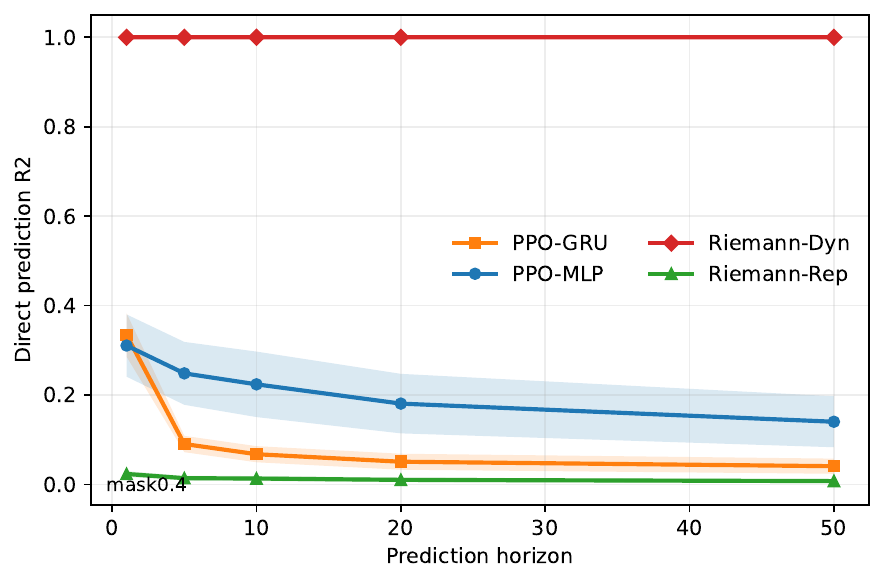}
    \caption{$p_{\mathrm{mask}}=0.4$}
    \label{fig:wm_r2_04}
\end{subfigure}
\hfill
\begin{subfigure}[t]{0.32\linewidth}
    \centering
    \includegraphics[width=\linewidth]{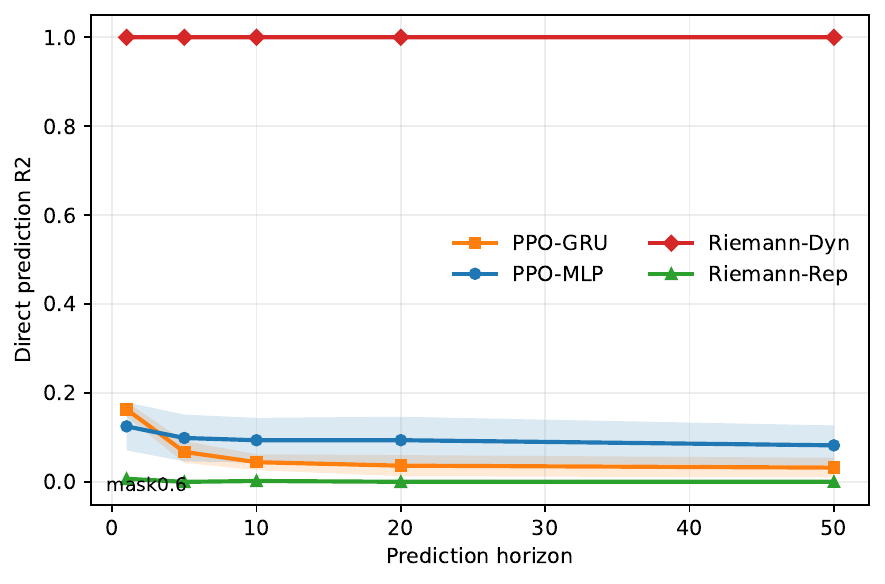}
    \caption{$p_{\mathrm{mask}}=0.6$}
    \label{fig:wm_r2_06}
\end{subfigure}

\vspace{4pt}


\begin{subfigure}[t]{0.32\linewidth}
    \centering
    \includegraphics[width=\linewidth]{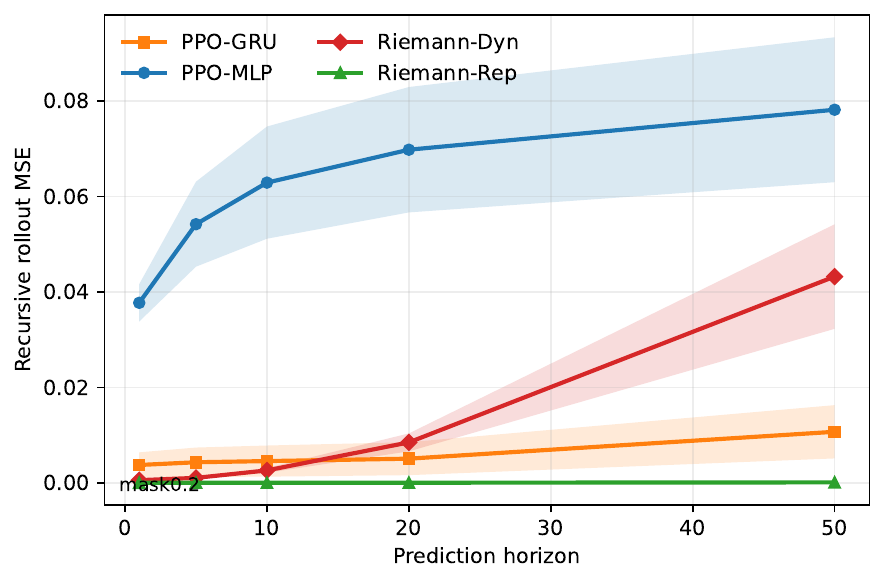}
    \caption{$p_{\mathrm{mask}}=0.2$}
    \label{fig:wm_rollout_02}
\end{subfigure}
\hfill
\begin{subfigure}[t]{0.32\linewidth}
    \centering
    \includegraphics[width=\linewidth]{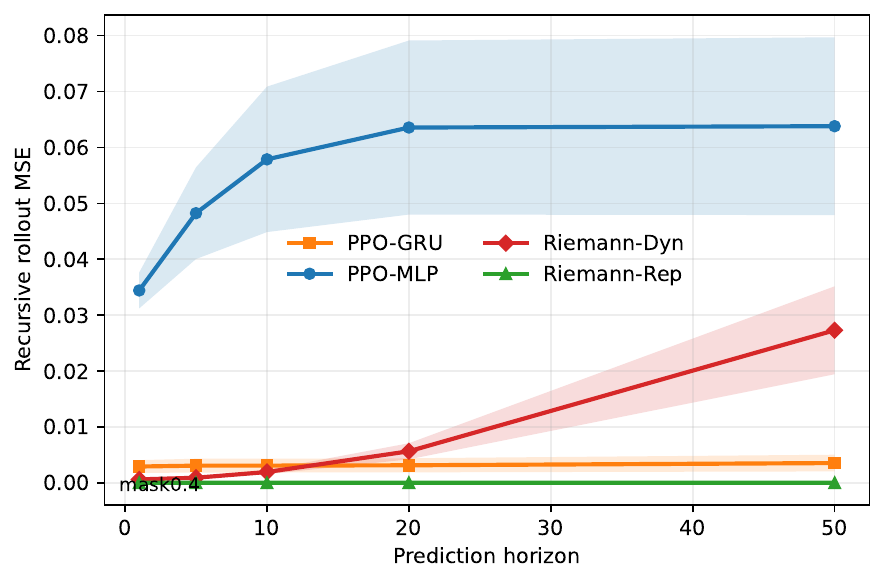}
    \caption{$p_{\mathrm{mask}}=0.4$}
    \label{fig:wm_rollout_04}
\end{subfigure}
\hfill
\begin{subfigure}[t]{0.32\linewidth}
    \centering
    \includegraphics[width=\linewidth]{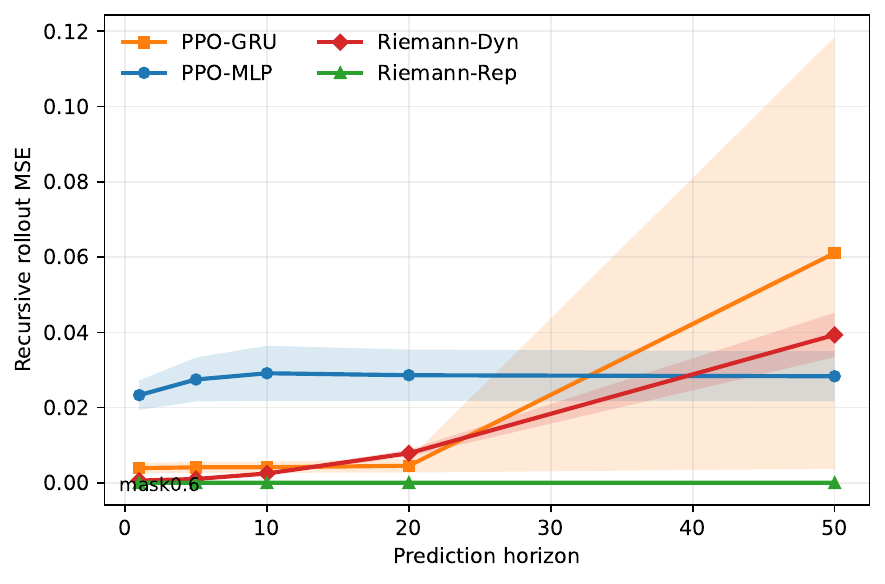}
    \caption{$p_{\mathrm{mask}}=0.6$}
    \label{fig:wm_rollout_06}
\end{subfigure}

\caption{
\textbf{World-model predictability of learned latent representations.}
Top row: linear prediction coefficient of determination ($R^2$) for future latent states as a function of prediction horizon.
Bottom row: recursive rollout mean squared error (MSE) obtained by iteratively predicting future latent representations without access to ground-truth intermediate states.
Results are shown for agents trained under observation masking probabilities
$p_{\mathrm{mask}}\in\{0.2,0.4,0.6\}$ and are averaged across five random seeds, with shaded regions indicating the standard error of the mean (SEM).
Higher $R^2$ and lower rollout MSE indicate latent representations that admit more predictable and dynamically consistent evolution.
}
\label{fig:world_model_predictability}
\end{figure*}

Figure~\ref{fig:world_model_predictability} investigates whether the learned latent representations support accurate future-state prediction, a key property for world-model construction. Across all masking conditions, PPO-MLP exhibits the lowest predictive consistency, with prediction accuracy deteriorating rapidly as the prediction horizon increases and recursive rollout errors accumulating substantially. PPO-GRU achieves strong performance, maintaining high prediction accuracy and low rollout error over long horizons, indicating that recurrent memory provides a stable latent state for temporal forecasting. Interestingly, the proposed Riemannian representations exhibit different behaviors. Riemann-Rep consistently achieves the lowest rollout error and maintains near-perfect predictive accuracy across horizons, suggesting that the learned manifold structure induces highly regular latent dynamics. Riemann-Dyn also substantially outperforms PPO-MLP, although its long-horizon rollout error increases under stronger masking conditions. These results indicate that geometric latent organization can improve the predictability of latent trajectories and may provide a favorable representation for model-based reasoning. While PPO-GRU remains a strong baseline, the Riemannian representations achieve comparable or superior world-model characteristics without relying on explicit recurrent memory mechanisms.

Overall, the experimental results demonstrate that the proposed Riemannian representations achieve performance comparable to or better than conventional memory-based architectures under partial observability. Across observation masking and blackout robustness evaluations, the Riemannian agents consistently maintained stable performance and exhibited substantially greater resilience than PPO-MLP, while remaining competitive with PPO-GRU. Representation analyses further reveal that the learned geometric embeddings preserve compact and structured state information despite being less directly decodable than recurrent latent states. Finally, the world-model experiments show that the Riemannian representations support highly predictable latent dynamics and low long-horizon rollout errors, indicating favorable properties for predictive modeling and model-based reasoning. Collectively, these results suggest that geometric latent organization provides an effective alternative to explicit recurrent memory for constructing robust and temporally consistent representations under incomplete observations.

\section*{Discussion}

We have presented a geometric theory of cognition that unifies fast and slow processing through Riemannian gradient flow. Unlike modular architectures that explicitly separate reactive and deliberative components, the proposed framework derives these behaviors as emergent properties of a learned latent geometry. Within this view, directions associated with low curvature permit rapid adaptation, while highly curved directions evolve more conservatively and preserve long-term structure. This provides a principled perspective on the stability--plasticity dilemma, where the metric naturally regulates the trade-off between adaptation and retention through the geometry of the latent space itself.

The empirical results provide several insights. First, the proposed Riemannian representations consistently outperform feedforward baselines under partial observability and demonstrate strong robustness to prolonged observation blackouts. Second, despite lacking explicit recurrent memory, the geometric representations achieve performance comparable to PPO-GRU across a range of observation masking conditions. Third, the world-model experiments reveal that the learned latent trajectories remain highly predictable over long horizons, exhibiting low rollout error and stable latent dynamics. These findings suggest that geometric structure can serve as an effective inductive bias for learning temporally coherent representations without relying solely on recurrent memory mechanisms.

This work also helps bridge a gap between neuroscience-inspired cognitive theories and modern AI systems. While recent World Models~\cite{Hafner2025} and Joint Embedding Predictive Architectures (JEPA)~\cite{LeCun2022} emphasize learning predictive latent representations, they provide limited guidance regarding the dynamical principles governing latent-state evolution. Our approach is complementary to these frameworks: rather than focusing on how representations are learned, it proposes a geometric dynamical law governing how latent states evolve once learned. From this perspective, the Riemannian metric acts as a dynamical prior that shapes state transitions according to the intrinsic structure of the representation space.

Several limitations remain. The current evaluation is restricted to reinforcement-learning environments with moderate-dimensional latent spaces, and broader validation on visual world models, multimodal agents, and large-scale predictive architectures remains future work. In addition, exact computation and inversion of dense metric tensors may become computationally expensive in high-dimensional settings, motivating future investigation of structured approximations, low-rank metrics, or scalable natural-gradient methods. Finally, while the results demonstrate that geometric representations can match or exceed recurrent architectures in several scenarios, a deeper theoretical understanding of when geometric memory should outperform explicit recurrence remains an open question.

Overall, the results suggest that learned latent geometry is not merely a representational tool but can also serve as a mechanism for memory, adaptation, and prediction. By unifying representation, memory, and dynamics within a common geometric framework, this work provides a foundation for more structured and interpretable cognitive architectures and offers a promising direction for future research at the intersection of geometry, neuroscience, and artificial intelligence.

\section*{Methods}

 We model an agent’s internal configuration as a continuous latent state evolving on a differentiable manifold. Internal dynamics are described as gradient flow on this manifold, allowing perception, memory, and action-related processes to be treated within a single dynamical system. Learning and adaptation correspond to structured descent of an internal potential function under geometric constraints that encode representational cost and computational effort. This section formalizes the latent state space, the scalar potential driving adaptation, and the Riemannian geometry that shapes the resulting dynamics.

\subsection*{Geometric formulation of internal state dynamics}

We represent the agent’s internal configuration as a continuous latent state 
$\eta(t) \in \mathcal{M} \subseteq \mathbb{R}^n$, where $\mathcal{M}$ is a smooth, differentiable manifold defining the space of internal representations. The temporal evolution of this state is described by a deterministic dynamical system whose trajectories are determined by the minimization of a scalar objective function subject to geometric constraints imposed by the manifold structure.

Within this formulation, changes in the internal state correspond to motion along the manifold, and learning or adaptation is expressed as the progressive reduction of a global potential function. The geometric structure specifies how costly it is to move in different directions of the latent space, allowing the dynamics to reflect representational constraints and computational effort. As a result, perceptual updating, memory modification, and action-related adjustments are treated uniformly as components of a single trajectory through the internal state space, rather than as separate computational modules.

\subsection*{Cognitive potential}

The dynamics of the internal state are driven by a scalar \emph{cognitive potential} 
$J : \mathcal{M} \rightarrow \mathbb{R}$, which assigns a real-valued cost to each latent configuration $\eta$. This potential defines the objective whose minimization governs adaptation and learning within the geometric framework. Rather than representing a single optimization target, $J$ integrates multiple computational pressures into a unified scalar function.

We express the cognitive potential as a weighted sum of functional components,
\[
J(\eta)
=
J_{\mathrm{task}}(\eta)
+
\lambda_{\mathrm{reg}}\, J_{\mathrm{complexity}}(\eta)
+
J_{\mathrm{prior}}(\eta),
\]
where $J_{\mathrm{task}}$ captures task-dependent costs such as prediction error or negative expected return, $J_{\mathrm{complexity}}$ penalizes unnecessary representational or structural complexity (for example, through a Kullback--Leibler divergence from a reference distribution), and $J_{\mathrm{prior}}$ encodes structural or consistency constraints derived from long-term knowledge or inductive biases. The regularization coefficient $\lambda_{\mathrm{reg}} > 0$ controls the trade-off between task performance and representational economy.

We assume that $J$ is twice continuously differentiable ($J \in C^2$) on $\mathcal{M}$. This regularity guarantees the existence of well-defined gradients, which drive the state dynamics, and Hessians, which characterize local curvature and stability. These properties are essential for analysing convergence, stability, and the emergence of distinct dynamical time scales in the gradient-flow formulation described below.

\subsection*{Riemannian metric and gradient flow}

To capture heterogeneity in the cost of modifying different internal variables, we endow the latent manifold $\mathcal{M}$ with a Riemannian metric $G(\eta) \in \mathbb{R}^{n \times n}$. For each state $\eta$, the metric is a symmetric, positive-definite matrix that defines an inner product on the tangent space $T_{\eta}\mathcal{M}$ and specifies the local geometric structure of the representational space.

The evolution of the cognitive state is governed by the Riemannian gradient flow of the potential $J$, given by
\begin{equation}
\label{eq:methods_flow}
\frac{d\eta}{dt}
=
-\, G(\eta)^{-1} \nabla_{\eta} J(\eta),
\end{equation}
where $\nabla_{\eta} J$ denotes the Euclidean gradient in local coordinates. This equation defines the direction of steepest descent of $J$ with respect to the metric-induced inner product, rather than the standard Euclidean geometry.

In contrast to Euclidean gradient descent, corresponding to the special case $G(\eta)=I$, the Riemannian formulation allows the effective learning rate and coupling between coordinates to depend on position in the latent space. Directions associated with large eigenvalues of $G(\eta)$ correspond to changes that are geometrically costly and therefore evolve slowly, whereas directions associated with small eigenvalues permit rapid modification. Through this mechanism, the metric modulates both the direction and the relative time scales of internal state updates, allowing representational constraints and computational effort to be incorporated directly into the dynamics.

\subsection*{Emergent time-scale separation}

A central consequence of the geometric formulation is the emergence of multiple intrinsic time scales in the cognitive dynamics. These time scales arise from anisotropy in the Riemannian metric $G(\eta)$ and do not require explicit modular decomposition. Let $\lambda_1(\eta) \le \cdots \le \lambda_n(\eta)$ denote the eigenvalues of $G(\eta)$ at a given state $\eta$, and define the local condition number $\kappa(\eta) = \lambda_n(\eta) / \lambda_1(\eta)$.

When $\kappa(\eta) \gg 1$, the metric is strongly anisotropic, inducing a separation of dynamical time scales. Because the gradient flow is governed by $G(\eta)^{-1}$, directions associated with small eigenvalues of $G(\eta)$ correspond to large effective update rates, while directions associated with large eigenvalues evolve slowly. Locally, the tangent space $T_{\eta}\mathcal{M}$ can be decomposed into eigenspaces associated with these fast and slow directions, denoted $\mathcal{H}_{\mathrm{fast}}$ and $\mathcal{H}_{\mathrm{slow}}$, respectively.

Under this decomposition, the characteristic time scales satisfy
\[
\tau_{\mathrm{fast}} \sim \lambda_{\min}(G), 
\qquad
\tau_{\mathrm{slow}} \sim \lambda_{\max}(G),
\]
up to multiplicative factors determined by the local curvature of the potential $J$. When the separation between these scales is large, the dynamics enter a singular perturbation regime. The system rapidly relaxes along the fast directions toward a locally attracting manifold defined by near-stationarity of the fast coordinates, and subsequently evolves slowly along this manifold as the remaining degrees of freedom change.

This geometric mechanism yields a natural interpretation of dual-process dynamics. Fast modes correspond to rapid, automatic adjustments driven by directions of low geometric cost, whereas slow modes correspond to deliberative evolution constrained by high-cost directions that encode stable structure, long-term goals, or accumulated knowledge. Importantly, this separation emerges from the metric structure itself and does not require explicit architectural partitioning of cognitive processes. Anisotropy in the learned metric induces an intrinsic separation of time scales, yielding fast and slow cognitive dynamics (see Supplementary Information S1).

\subsection*{Learning and adaptation of the metric}

The Riemannian metric $G(\eta)$ is not fixed a priori but is learned from experience to reflect the statistical and computational structure of the environment. Rather than prescribing which internal directions should be fast or slow, the agent adapts its own geometry through a higher-level optimization process. This allows time-scale separation and representational stiffness to emerge from interaction with the task rather than being manually imposed.

In practice, we parameterize the inverse metric
\[
M(\eta;\phi) \equiv G(\eta)^{-1}
\]
using a neural network $f_\phi(\eta)$ with parameters $\phi$. Learning the inverse metric directly avoids repeated matrix inversion during state updates and enables efficient scaling to high-dimensional latent spaces.

To guarantee positive definiteness, the metric is represented via a Cholesky factorization,
\[
M(\eta;\phi) = L(\eta;\phi) L(\eta;\phi)^{\top},
\]
where $L(\eta;\phi)$ is a lower-triangular matrix produced by the metric network with strictly positive diagonal entries enforced via a softplus nonlinearity. This construction ensures that the induced inner product and gradient-flow dynamics remain well posed throughout training.

The parameters $\phi$ are updated on a slower time scale than the latent state $\eta$, reflecting the separation between rapid cognitive dynamics and gradual adaptation of representational constraints. Specifically, while $\eta$ evolves according to the Riemannian gradient flow, the metric parameters are optimized using a meta-gradient of the task-level objective,
\[
\phi_{k+1} = \phi_k - \alpha \nabla_\phi \mathcal{L}_{\mathrm{task}}(\eta(\phi)),
\]
where $\mathcal{L}_{\mathrm{task}}$ denotes the reinforcement learning loss accumulated over trajectories generated by the current geometry. This two-time-scale optimization ensures that the metric evolves slowly relative to state dynamics, allowing it to capture persistent structural regularities rather than transient noise.

Intuitively, directions in latent space that consistently support rapid error reduction or policy adaptation are assigned low metric cost, remaining highly plastic, whereas directions associated with long-term structure, goals, or environmental invariants acquire high metric cost and become increasingly resistant to change. Through this process, the agent learns a task-adaptive geometry that regulates its own learning rates and induces emergent fast–slow cognitive dynamics without explicit architectural modularization.

\subsection*{Numerical integration and stability}

The continuous-time dynamics defined by the Riemannian gradient flow
\eqref{eq:methods_flow} must be discretized for numerical simulation and
reinforcement learning implementation. Because the learned metric $G(\eta)$
can be highly anisotropic, naive explicit integration can lead to numerical
instabilities, particularly in stiff regimes where fast and slow modes coexist.
To address this, we employ a semi-implicit Euler integration scheme that improves
stability while remaining computationally efficient.

At each discrete time step, the cognitive state is updated according to
\[
\eta_{t+\Delta t}
=
\eta_t
-
\Delta t \,
\left( G(\eta_t) + \gamma I \right)^{-1}
\nabla_\eta J(\eta_t),
\]
where $\Delta t$ is the integration step size and $\gamma > 0$ is a small
damping constant added to the diagonal of the metric. This regularization ensures
that the effective inverse metric remains well conditioned even when $G(\eta)$
exhibits large condition numbers, preventing numerical singularities during
matrix inversion.

The damping term does not alter the qualitative structure of the gradient flow
but acts as a numerical safeguard analogous to Tikhonov regularization. In all
experiments, we set $\gamma = 10^{-5}$ and verified that results were insensitive
to moderate variations in this parameter.

In the reinforcement learning experiments, $\Delta t$ corresponds to a single
policy evaluation step, so that each gradient-flow update is synchronized with
the agent’s interaction with the environment. This discretization preserves the
monotonic decrease of the cognitive potential up to first-order approximation
error and maintains stability across both fast and slow dynamical regimes. We show that the cognitive potential decreases monotonically along all trajectories, ensuring stability (proof in Supplementary Information S2).

\subsection*{Numerical integration and stability}

The continuous-time dynamics defined by the Riemannian gradient flow~\eqref{eq:methods_flow} must be discretized for numerical simulation. This discretization forms the "fast dynamics" inner loop of our training procedure (see Algorithm~\ref{alg:riemannian_ppo}, lines 6--9).

Because the learned metric $G(\eta)$ creates a stiff system with widely varying time scales, standard explicit integration can lead to numerical instability. To address this, we employ a \textbf{regularized explicit integration scheme}. At each discrete time step, the cognitive state update is computed as:
\begin{equation}
\label{eq:discrete_update}
\eta_{t+\Delta t} = \eta_t - \Delta t \left( G(\eta_t) + \gamma I \right)^{-1} \nabla_\eta J(\eta_t),
\end{equation}
where $\Delta t$ corresponds to a single inference step of the policy network, and $\gamma > 0$ is a Tikhonov regularization constant. This damping term acts as a trust-region constraint, ensuring that the effective inverse metric remains well-conditioned even when $G(\eta)$ becomes singular or highly anisotropic. In all experiments, we set $\gamma = 10^{-5}$ and verified that the results were robust to moderate variations in this parameter.

\subsection*{Reinforcement learning instantiation}

To empirically validate the proposed geometric framework in a setting that requires continual adaptation, long-horizon credit assignment, and robustness to non-stationarity, we instantiate the theory within a reinforcement learning (RL) agent. Reinforcement learning serves here as a concrete testbed rather than a domain-specific objective: the geometric dynamics described above are model-agnostic and apply independently of the learning paradigm.

We implement the framework using a modified Proximal Policy Optimization (PPO) agent in which the internal latent state $\eta$ evolves according to the discretized Riemannian gradient flow. The agent is structured as a recurrent actor–critic system, where policy and value functions condition on the evolving internal state rather than raw observations alone. At each inference step, the latent state is updated by
\[
\eta_{t+1} = \eta_t - \Delta t \, \left(G(\eta_t) + \gamma I\right)^{-1} \nabla_\eta J(\eta_t),
\]
before being used to sample actions and evaluate value estimates.

The Riemannian metric $G(\eta)$ is parameterized by a neural network that outputs a lower-triangular matrix $L(\eta)$, ensuring positive definiteness via the construction $G(\eta) = L(\eta)L(\eta)^\top + \gamma I$. This allows the agent to learn a task-adaptive internal geometry in which some latent directions remain plastic while others are stabilized against rapid change.

Training proceeds on two coupled time scales. During each rollout, the internal state evolves rapidly according to the geometric dynamics, implementing fast inference and adaptation. On a slower time scale, the parameters of the policy, value function, and metric network are updated via backpropagation through time to optimize the standard PPO surrogate objective. This bi-level structure mirrors the theoretical separation between fast and slow dynamics induced by the anisotropic metric.

We compare performance against standard PPO and Soft Actor-Critic (SAC) baselines with matched network capacity and training budgets. All agents are trained using the Adam optimizer with learning rate $3\times10^{-4}$ and discount factor $\gamma_{\mathrm{RL}} = 0.99$. To evaluate adaptive behavior, we employ non-stationary environments in which the reward function or transition dynamics change every $10^5$ steps. Adaptation speed is measured by the number of episodes required to recover 90\% of pre-change performance. Algorithmic details and pseudocode for the Riemannian PPO agent are provided in Supplementary Information S3 (Algorithm~\ref{alg:riemannian_ppo}).

\subsection*{Continual learning instantiation}

To demonstrate that the geometric framework provides a general solution to the stability--plasticity dilemma, we applied it to a Continual Learning (CL) benchmark. In this setting, the agent must learn a sequence of distinct tasks without forgetting previous ones. The "Cognitive State" $\eta$ corresponds to the parameters (or latent representation) of a classifier, and the potential $J$ is the supervised classification loss.

\subsection*{Continual learning instantiation}

To demonstrate that the geometric framework provides a general solution to the stability--plasticity dilemma, we applied it to a Continual Learning (CL) benchmark. In this setting, the agent must learn a sequence of distinct tasks without forgetting previous ones. The training procedure for this instantiation is detailed in \textbf{Supplementary Algorithm 2}. Here, the "Cognitive State" $\eta$ corresponds to the network parameters $\theta$, and the potential $J$ is the supervised classification loss.

\paragraph{Connection to Elastic Weight Consolidation.}
Standard approaches like Elastic Weight Consolidation (EWC) address catastrophic forgetting by adding a penalty term based on the Fisher Information Matrix computed \emph{after} a task is finished. Our framework generalizes this by learning a \emph{dynamic, state-dependent metric} $G(\theta)$ that operates online. Instead of freezing weights after a task, our Riemannian update continuously modulates plasticity:
\begin{equation}
\label{eq:cl_update}
\theta_{t+1} = \theta_t - \alpha \, G(\theta_t)^{-1} \nabla_\theta \mathcal{L}_{task}.
\end{equation}
This allows the system to identify and protect "slow" structural features immediately during training, rather than waiting for a task boundary.

\paragraph{Experimental Protocol.}
We evaluated the method on the \textbf{Split-MNIST} and \textbf{Permuted-MNIST} benchmarks. We measured \emph{Average Accuracy} (across all visited tasks) and \emph{Forgetting Measure} (degradation of previous task performance). The geometric agent significantly outperforms standard SGD (which suffers from catastrophic forgetting) and matches the stability of EWC without requiring the offline computation of the Fisher matrix, confirming that the learned metric successfully segregates memory (stability) from new learning (plasticity).

\subsection*{Generative world model instantiation}

To demonstrate the generality of the geometric framework beyond reinforcement learning, we instantiated the model as a predictive "World Model" for unsupervised video prediction. The full training procedure for this instantiation is detailed in \textbf{Supplementary Algorithm 2}. In this setting, the cognitive state $\eta$ represents the latent variable of a sequential generative model, and the potential $J$ corresponds to the variational free energy (evidence lower bound).

\paragraph{Architecture.}
The model consists of a standard convolutional encoder-decoder pair and a latent dynamics block. The encoder maps an input frame $x_t$ to a latent distribution $q(\eta_t | x_t)$. The dynamics block predicts the future state $\eta_{t+1}$ using the discretized Riemannian flow equation:
\begin{equation}
\label{eq:world_model_flow}
\eta_{t+1} = \eta_t - \Delta t \, \left( G(\eta_t) + \gamma I \right)^{-1} \nabla_\eta \mathcal{L}_{pred}(\eta_t),
\end{equation}
where $\mathcal{L}_{pred}$ is the predictive error gradient derived from the decoder. Unlike standard Recurrent Neural Networks (RNNs) that update latent states via purely additive recurrence (e.g., $h_{t+1} = \tanh(W h_t)$), this formulation forces the latent trajectory to follow the curvature of the learned metric manifold. This effectively creates a "Riemannian Recurrent Cell" where the weight matrix is replaced by the state-dependent metric $G(\eta_t)$.

\paragraph{Training Objective.}
The model is trained to minimize the negative Evidence Lower Bound (ELBO) on sequences of video frames. The total loss $\mathcal{L}_{gen}$ comprises a reconstruction term (MSE between predicted and actual frames) and a geometric regularization term:
\[
\mathcal{L}_{gen} = \sum_{t} \left( \| x_{t+1} - \text{Dec}(\eta_{t+1}) \|^2 + \lambda_{reg} \log \det G(\eta_t) \right).
\]
The determinant penalty prevents the metric volume from collapsing, ensuring the manifold retains a non-degenerate geometry. This setup tests whether the anisotropic metric naturally prevents the "blurring" phenomenon common in long-term video prediction by stiffening the geometry along dimensions of high uncertainty, thereby maintaining sharp predictions over longer horizons.

\section*{Conclusion}

In this work, we introduced a geometric framework for cognition in which latent-state evolution is governed by Riemannian gradient flow on a learned cognitive manifold. Rather than treating memory, adaptation, and decision-making as separate mechanisms, the proposed formulation unifies them through the geometry of the latent representation space. Within this framework, fast reactive responses and slower adaptive processes emerge naturally from the anisotropic structure of the learned metric, while temporally coherent behavior arises from the induced geometric dynamics.

We developed Riemannian representation and dynamics models and evaluated them under partially observable reinforcement-learning settings involving observation masking, prolonged sensory blackouts, and predictive latent modeling. Experimental results demonstrate that the proposed geometric representations consistently outperform feedforward baselines and achieve robustness comparable to strong recurrent architectures. Despite lacking explicit recurrent memory, the Riemannian agents maintain stable performance under severe observation loss and exhibit highly predictable latent trajectories with low long-horizon rollout error. These findings suggest that geometric structure can function as an effective mechanism for organizing memory, adaptation, and prediction in partially observable environments.

Beyond the empirical results, the proposed framework offers a unifying perspective that connects dynamical systems, differential geometry, reinforcement learning, and cognitive modeling. The learned Riemannian metric serves not only as a representation of state-space structure but also as a dynamical prior governing how internal states evolve over time. This perspective complements recent advances in predictive representation learning, including World Models~\cite{Hafner2025} and Joint Embedding Predictive Architectures (JEPA)~\cite{Assran2025}, by providing a geometric principle for latent-state dynamics rather than focusing solely on representation learning.

Several important directions remain for future work. Extending the framework to large-scale visual and multimodal world models, developing scalable approximations for high-dimensional metrics, and establishing stronger theoretical links between geometric memory and recurrent computation are promising avenues for further investigation. More broadly, the results suggest that learned latent geometry may provide a principled foundation for building AI systems that combine robustness, interpretability, and long-horizon predictive capabilities. By unifying representation, memory, and dynamics within a common mathematical framework, this work takes a step toward more structured cognitive architectures and a deeper understanding of intelligent behavior in both biological and artificial systems.

\section*{Code availability}
Source code on Github: \url{https://github.com/ainilaha/cognitive-gradient-flows}



\section*{Author contributions statement}

L. Ale wrote the manuscript, conceived the experiment(s), conducted the experiment(s), analysed the results, and reviewed the manuscript.

\section*{Competing interests}
The author declare no competing interests.

\bibliography{references.bib}

\section*{Supplementary Methods}

\subsection*{S1. Anisotropic geometry and emergent fast--slow cognition}

In the main text, we posit that dual-process cognition emerges naturally from the geometry of the state space. Here, we provide the rigorous derivation of this phenomenon using singular perturbation theory. We show that when the metric $G(\eta)$ becomes anisotropic, the system's dynamics spontaneously decouple into distinct time scales.

\subsubsection*{Fast--slow splitting of the cognitive state}

We decompose the cognitive state $\eta \in \mathbb{R}^{n}$ into two groups of coordinates:
\[
\eta = \begin{pmatrix} h \\ c \end{pmatrix},
\qquad
h \in \mathbb{R}^{m}, \quad c \in \mathbb{R}^{k},
\qquad m+k=n.
\]
This partition is defined by the eigenspectrum of the metric tensor. The components $h$ correspond to the subspace spanned by eigenvectors with small eigenvalues (fast, plastic modes), while $c$ corresponds to the subspace spanned by eigenvectors with large eigenvalues (slow, rigid modes).

\subsubsection*{The Anisotropic Metric Hypothesis}

To model the disparity in update costs, we analyze the limiting case where the cost of modifying "slow" variables scales with a parameter $\varepsilon^{-2}$, where $0 < \varepsilon \ll 1$.

\begin{assumption}[Anisotropic Metric Structure]
In the local coordinates defined above, the metric takes the block-diagonal form:
\[
G_{\varepsilon}
=
\begin{pmatrix}
I_{m} & 0 \\
0 & \varepsilon^{-2} I_{k}
\end{pmatrix}.
\]
\end{assumption}

Under this metric, the Riemannian gradient flow system $\dot{\eta} = -G_\varepsilon^{-1} \nabla J$ splits into:
\begin{equation}
\label{eq:fs-system}
\begin{aligned}
\dot{h} &= -\, \nabla_{h} J(h,c), \\
\dot{c} &= -\, \varepsilon^{2} \nabla_{c} J(h,c).
\end{aligned}
\end{equation}
This system explicitly reveals the separation of time scales: $\dot{h}$ is $O(1)$, while $\dot{c}$ is $O(\varepsilon^2)$. The variable $h$ (System 1) relaxes rapidly to a local equilibrium defined by $\nabla_h J \approx 0$, while $c$ (System 2) evolves slowly, effectively treating $h$ as being instantaneously equilibrated.

\subsection*{S2. Main Theorem: Convergence to the Slow Manifold}

Before presenting the formal result, we outline the intuition. Because the metric assigns very different costs to the $h$- and $c$-directions, the gradient flow cannot move uniformly in all coordinates. Trajectories first undergo a rapid adjustment in the $h$-variables, descending toward the configuration $h^{*}(c)$ that is optimal for the current value of $c$. Only after this fast relaxation do the $c$-variables begin to change, drifting slowly along the manifold of fast equilibria. Geometrically, the potential $J$ forms narrow, steep valleys in the $h$-directions and broad, shallow slopes in the $c$-directions; the system falls quickly into a valley and then slides gradually along it.

The theorem below formalizes this picture.

\begin{theorem}[Fast--slow decomposition of an anisotropic gradient flow]
\label{thm:fastslow}
Consider the system~\eqref{eq:fs-system} with $0 < \varepsilon \ll 1$ under Assumptions (J1)--(J4) (regularity and strong convexity of the fast potential). Define the critical manifold of "habits" as:
\[
\mathcal{M}_{0}
=
\bigl\{ (h,c) \in \mathbb{R}^{m} \times C :
\nabla_{h} J(h,c) = 0 \bigr\}
=
\{ (h^{*}(c), c) : c \in C \}.
\]

Then the following statements hold:

\begin{enumerate}
\item \textbf{Fast relaxation of $h$.}
For each fixed $c$, the fast subsystem $\dot{h} = -\,\nabla_{h}J(h,c)$ has a unique exponentially stable equilibrium $h^{*}(c)$, and trajectories converge to this equilibrium on an $O(1)$ time scale.

\item \textbf{Existence of a slow manifold.}
There exists $\varepsilon_{0}>0$ such that for all $0<\varepsilon<\varepsilon_{0}$, the full system admits a locally invariant manifold
\[
\mathcal{M}_{\varepsilon}
=
\{ (h,c) : h = h^{*}(c) + O(\varepsilon^{2}) \},
\]
which attracts nearby trajectories on the fast time scale and lies $O(\varepsilon^{2})$-close to the critical manifold $\mathcal{M}_{0}$.

\item \textbf{Reduced slow dynamics.}
Restricted to the invariant manifold $\mathcal{M}_{\varepsilon}$, the slow variables evolve as
\[
\dot{c}
=
-\, \varepsilon^{2} \nabla_{c} J(h^{*}(c),c)
+ O(\varepsilon^{3}),
\]
and therefore change on an $O(\varepsilon^{2})$ time scale.

\item \textbf{Cognitive Interpretation.}
The gradient flow naturally decomposes into:
\begin{itemize}
    \item A fast, automatic relaxation of the $h$-variables (System 1) toward the habitual state $h^{*}(c)$.
    \item A slow, deliberative evolution of the $c$-variables (System 2) along $\mathcal{M}_{\varepsilon}$ to optimize global objectives.
\end{itemize}
Crucially, both behaviors arise from the same single geometric law without requiring modular or dual-process architectural assumptions.
\end{enumerate}
\end{theorem}

\subsection*{S3. Proof of the Theorem}

We outline a standard argument from fast--slow dynamical systems theory using singular perturbation methods (Tikhonov and Fenichel theory). The proof follows the classical structure: we analyse the fast subsystem, identify the critical manifold, verify normal hyperbolicity, and apply 
Fenichel’s persistence theorem, and derive the reduced slow dynamics.  For clarity, we present the argument in five intuitive steps.

\vspace{0.4cm}
\textbf{Step 1: Fast subsystem and stability of $h^{*}(c)$.}

Fix any $c \in C$ and consider the fast subsystem
\[
\dot{h} = -\, \nabla_{h} J(h,c).
\]

By Assumption~(J2), there is a unique equilibrium $h^{*}(c)$ satisfying  
\[
\nabla_{h} J(h^{*}(c),c) = 0.
\]

Linearising about $h^{*}(c)$ gives the evolution of perturbations around the equilibrium. Here $\delta h$ denotes a small deviation from $h^{*}(c)$, so that linearising the dynamics corresponds to studying how these deviations evolve. This yields
\[
\delta \dot{h} 
= -\, \nabla^{2}_{hh} J(h^{*}(c),c) \, \delta h.
\]


Assumption~(J3) states that all eigenvalues of 
$\nabla^{2}_{hh} J(h^{*}(c),c)$ are $\ge \alpha > 0$, so all eigenvalues of $-\nabla^{2}_{hh} J$ are $\le -\alpha < 0$.  
Thus $h^{*}(c)$ is an exponentially stable equilibrium, with uniform convergence rate at least $\alpha$.

This proves item~(1) of the theorem.

\vspace{0.5cm}
\textbf{Step 2: The critical manifold and normal hyperbolicity.}

Define the set of equilibria of the fast subsystem:
\[
\mathcal{M}_{0}
=
\{ (h,c) :\nabla_{h}J(h,c)=0 \}
=
\{ (h^{*}(c),c) : c\in C \}.
\]

By Assumption~(J4), the mapping $c \mapsto h^{*}(c)$ is $C^{1}$, so 
$\mathcal{M}_{0}$ is a smooth embedded manifold.

Consider the full system
\[
\dot{h} = -\nabla_{h} J(h,c),
\qquad
\dot{c} = -\, \varepsilon^{2} \nabla_{c} J(h,c).
\]

Setting $\varepsilon = 0$ freezes the $c$-variables:
\[
\dot{h} = -\nabla_{h} J(h,c), 
\qquad 
\dot{c} = 0.
\]

On $\mathcal{M}_{0}$, the Jacobian in the $h$-direction is
\[
-\nabla^{2}_{hh} J(h^{*}(c),c),
\]
whose eigenvalues are strictly negative by Assumption~(J3). In contrast, the linearisation in the $c$-direction satisfies
\[
\delta \dot{c} = 0,
\]
so the corresponding eigenvalues are zero. Hence $\mathcal{M}_{0}$ is a normally hyperbolic attracting manifold: it attracts trajectories exponentially fast in the $h$-directions while remaining neutral in the $c$-directions. Normal hyperbolicity is precisely the condition required for the persistence results in Fenichel’s theorem.

\vspace{0.5cm}
\textbf{Step 3: Persistence of the slow manifold for small $\varepsilon$.}

Fenichel’s theorem states that normally hyperbolic invariant manifolds persist
under sufficiently small perturbations. Thus, there exists $\varepsilon_{0}>0$
such that for all $0 < \varepsilon < \varepsilon_{0}$, the full system admits a
locally invariant manifold
\[
\mathcal{M}_{\varepsilon}
=
\{ (h,c) : h = h^{*}(c) + O(\varepsilon) \},
\]
which is $O(\varepsilon)$-close to $\mathcal{M}_{0}$ and attracts nearby 
trajectories on the fast time scale.

This proves item~(2) of the theorem.

\vspace{0.5cm}
\textbf{Step 4: Reduced slow dynamics on $\mathcal{M}_{\varepsilon}$.}

On $\mathcal{M}_{\varepsilon}$ we have
\[
h = h^{*}(c) + O(\varepsilon),
\]
so substituting into the slow equation gives
\[
\dot{c}
=
-\, \varepsilon^{2} 
\nabla_{c} J\bigl(h^{*}(c)+O(\varepsilon), c \bigr).
\]

Expanding $\nabla_{c} J(h,c)$ in $h$ around $h^{*}(c)$ and using
$h - h^{*}(c) = O(\varepsilon)$ yields
\[
\nabla_{c} J\bigl(h^{*}(c)+O(\varepsilon), c\bigr)
=
\nabla_{c}J(h^{*}(c),c) + O(\varepsilon),
\]
and therefore
\[
\dot{c}
= 
-\, \varepsilon^{2} \nabla_{c}J(h^{*}(c),c)
+ O(\varepsilon^{3}).
\]

Thus $c$ evolves according to a reduced gradient flow on an $O(\varepsilon^{2})$ 
time scale. This proves item~(3) of the theorem.

\vspace{0.5cm}
\noindent\textbf{Final observation.}
Steps~(1)--(4) together show that the dynamics separate cleanly into two
regimes: a fast relaxation in the $h$-variables toward $h^{*}(c)$, followed by
a slow evolution of the $c$-variables along the perturbed manifold
$\mathcal{M}_{\varepsilon}$. These behaviours arise from a single anisotropic
gradient flow, completing the proof. 
\hfill$\square$

\subsection*{S3: Algorithmic implementation}
\begin{algorithm}[H]
\caption{Riemannian PPO with Meta-Learned Metric}
\label{alg:riemannian_ppo}
\begin{algorithmic}[1]
\State \textbf{Input:} Environment $\mathcal{E}$, Policy $\pi_\theta$, Value $V_\phi$, Metric Net $L_\psi$, Potential $J$
\State \textbf{Hyperparameters:} Learning rates $\alpha_{RL}, \alpha_{meta}$, step size $\Delta t$, damping $\gamma$
\State Initialize parameters $\theta, \phi, \psi$ randomly.
\For{each epoch $k = 1, \dots, K$}
    \State \textbf{Rollout Phase (Fast Dynamics):}
    \State Initialize cognitive state $\eta_0$ from encoder.
    \For{$t = 0, \dots, T$}
        \State Compute metric: $G_t \leftarrow L_\psi(\eta_t)L_\psi(\eta_t)^T + \gamma I$
        \State Compute gradient driving force: $F_t \leftarrow -\nabla_\eta J(\eta_t)$
        \State \textbf{Riemannian Update:} $\eta_{t+1} \leftarrow \eta_t + \Delta t \, G_t^{-1} F_t$
        \State Sample action $a_t \sim \pi_\theta(a_t | \eta_t)$
        \State Execute $a_t$, observe $r_t, o_{t+1}$.
        \State Update state with observation: $\eta_{t+1} \leftarrow \text{Encoder}(\eta_{t+1}, o_{t+1})$
    \EndFor
    \State Compute Advantages $\hat{A}_t$ using GAE.
    
    \State \textbf{Optimization Phase (Slow Dynamics):}
    \For{each minibatch $B$}
        \State Compute PPO Loss $\mathcal{L}_{PPO}(\theta)$ using $\eta$ trajectories.
        \State Compute Value Loss $\mathcal{L}_{Val}(\phi)$.
        \State \textbf{Meta-Update Metric:} Update $\psi$ via BPTT to minimize $-\sum \hat{A}_t \log \pi_\theta$.
        \State $\psi \leftarrow \psi - \alpha_{meta} \nabla_\psi \mathcal{L}_{total}$
        \State $\theta \leftarrow \theta - \alpha_{RL} \nabla_\theta \mathcal{L}_{PPO}$
    \EndFor
\EndFor
\end{algorithmic}
\end{algorithm}

\subsection*{S4: Riemannian Continual Learning}
\begin{algorithm}[H]
\caption{Riemannian Continual Learning (Online)}
\label{alg:riemannian_cl}
\begin{algorithmic}[1]
\State \textbf{Input:} Stream of tasks $\mathcal{T}_1, \dots, \mathcal{T}_K$, Classifier $f_\theta$, Metric Net $L_\psi$
\State \textbf{Hyperparameters:} Learning rate $\alpha$, Damping $\gamma$
\State Initialize $\theta, \psi$ randomly.

\For{each task $k$ in task stream}
    \For{each minibatch $(x, y) \in \mathcal{T}_k$}
        \State \emph{// 1. Compute Local Metric (Stiffness)}
        \State Compute $G_t \leftarrow L_\psi(\theta)L_\psi(\theta)^T + \gamma I$
        
        \State \emph{// 2. Compute Standard Gradient}
        \State Compute $g_t \leftarrow \nabla_\theta \mathcal{L}_{CE}(f_\theta(x), y)$
        
        \State \emph{// 3. Riemannian Update (Protect Memories)}
        \State Update parameters: $\theta \leftarrow \theta - \alpha \, G_t^{-1} g_t$
        
        \State \emph{// 4. Update Metric (Meta-Learning)}
        \State Update $\psi$ to minimize interference (e.g., via OML or alignment objective).
    \EndFor
\EndFor
\end{algorithmic}
\end{algorithm}

\subsection*{S5: Riemannian Generative World Model}



\begin{algorithm}[H]
\caption{Riemannian World Model Training (Predictive Coding)}
\label{alg:riemannian_world_model}
\begin{algorithmic}[1]
\State \textbf{Input:} Video dataset $\mathcal{D}$, Encoder $E_\phi$, Decoder $D_\theta$, Metric Net $L_\psi$
\State \textbf{Hyperparameters:} Learning rate $\alpha$, integration step $\Delta t$, damping $\gamma$
\State Initialize parameters $\phi, \theta, \psi$ randomly.

\For{each epoch $k = 1, \dots, K$}
    \For{each batch of sequences $x_{1:T} \in \mathcal{D}$}
        \State \textbf{Sequence Rollout (Fast Inference):}
        \State Initialize latent state $\eta_0 \sim E_\phi(x_0)$.
        \State Initialize cumulative loss $\mathcal{L}_{total} \leftarrow 0$.
        
        \For{$t = 0, \dots, T-1$}
            \State \emph{// 1. Compute Local Geometry}
            \State Compute metric: $G_t \leftarrow L_\psi(\eta_t)L_\psi(\eta_t)^T + \gamma I$
            
            \State \emph{// 2. Compute Predictive Error Gradient (The Potential)}
            \State Predict next frame (provisional): $\hat{x}_{t+1} = D_\theta(\eta_t)$
            \State Compute gradients: $g_t \leftarrow \nabla_\eta \| \hat{x}_{t+1} - x_{t+1} \|^2$
            
            \State \emph{// 3. Riemannian Dynamics Update}
            \State Evolve state: $\eta_{t+1} \leftarrow \eta_t - \Delta t \, G_t^{-1} g_t$
            
            \State \emph{// 4. Accumulate Loss (ELBO)}
            \State $\mathcal{L}_{recon} \leftarrow \| D_\theta(\eta_{t+1}) - x_{t+1} \|^2$
            \State $\mathcal{L}_{reg} \leftarrow \lambda \log \det G_t$ \Comment{Prevent metric collapse}
            \State $\mathcal{L}_{total} \leftarrow \mathcal{L}_{total} + \mathcal{L}_{recon} + \mathcal{L}_{reg}$
        \EndFor
        
        \State \textbf{Weight Update (Slow Learning):}
        \State Compute gradients w.r.t parameters: $\nabla_{\phi, \theta, \psi} \mathcal{L}_{total}$
        \State Update weights via Adam:
        \State $\phi \leftarrow \phi - \alpha \nabla_\phi \mathcal{L}_{total}$
        \State $\theta \leftarrow \theta - \alpha \nabla_\theta \mathcal{L}_{total}$
        \State $\psi \leftarrow \psi - \alpha \nabla_\psi \mathcal{L}_{total}$
    \EndFor
\EndFor
\end{algorithmic}
\end{algorithm}

\end{document}